\documentclass[11pt]{article}

\usepackage[preprint]{acl}

\usepackage{times}
\usepackage{latexsym}

\usepackage[T1]{fontenc}

\usepackage[utf8]{inputenc}

\usepackage{microtype}

\usepackage{inconsolata}

\usepackage{graphicx}

\usepackage{balance} 
\usepackage{pifont}
\usepackage{algorithm}
\usepackage{algpseudocode}
\usepackage{booktabs}
\usepackage{arydshln}
\usepackage{subfigure}
\usepackage{graphicx}
\usepackage{tcolorbox}
\usepackage{array}
\usepackage{multirow}
\usepackage{fontawesome}
\usepackage{xcolor}
\usepackage{multicol}
\usepackage{enumitem}
\usepackage{amsmath}
\usepackage[table]{xcolor}
\usepackage{tabularx}

\title{TeamLLM: A Human-Like Team-Oriented Collaboration Framework for Multi-Step Contextualized Tasks}
\author{
 \textbf{Xiangyu Wang\textsuperscript{1}},
 \textbf{Jin Wu\textsuperscript{2,3,4}},
 \textbf{Haoran Shi\textsuperscript{1}},
 \textbf{Wei Xia\textsuperscript{1}},
 \\
 \textbf{Jiarui Yu\textsuperscript{4}},
 \textbf{Chanjin Zheng\textsuperscript{2,3}},
\\
 \textsuperscript{1}Department of Educational Psychology, East China Normal University \\
 \textsuperscript{2}Lab of Artificial Intelligence for Education, East China Normal University \\
 \textsuperscript{3}Shanghai Institute of Artificial Intelligence for Education, East China Normal University \\
 \textsuperscript{4}School of Computer Science and Technology, East China Normal University \\
\href{mailto:email@domain}{51274118009@stu.ecnu.edu.cn}, \href{mailto:email@domain}{chjzheng@dep.ecnu.edu.cn}
 }

\begin{document}
\maketitle
\begin{abstract}
Recently, multi-Large Language Model (LLM) frameworks have been proposed to solve contextualized tasks. However, these frameworks do not explicitly emulate human team role division, which may lead to a single perspective, thereby weakening performance on multi-step contextualized tasks. To address this issue, we propose \textbf{TeamLLM}, a human-like Team-Oriented Multi-LLM Collaboration Framework. TeamLLM adopts \textbf{four team roles} with distinct division and employs a \textbf{three-phase multi-LLM collaboration} for multi-step contextualized tasks. To evaluate the effectiveness of TeamLLM on multi-step contextualized tasks, we propose \textbf{C}ontextually-\textbf{G}rounded and \textbf{P}rocedurally-\textbf{S}tructured tasks (\textbf{CGPST}) and construct the CGPST benchmark. This benchmark has four core features: contextual grounding, procedural structure, process-oriented evaluation and multi-dimensional assessment. We evaluate ten popular LLMs on CGPST at overall-level, step-level, and dimension-level. Results show that TeamLLM substantially improves performance on CGPST. We release the benchmark with scenarios, full-process responses and human scores from ten LLMs. The code and data are available at \url{https://anonymous.4open.science/r/TeamLLM-anonymous-C50E/}\footnote{This is an anonymous GitHub link containing partial data, code, and resources. The full GitHub repository will be made publicly available after acceptance.}.
\end{abstract}
\section{Introduction}
\begin{figure}[t]  
  \centering
  \includegraphics[width=0.95\columnwidth]{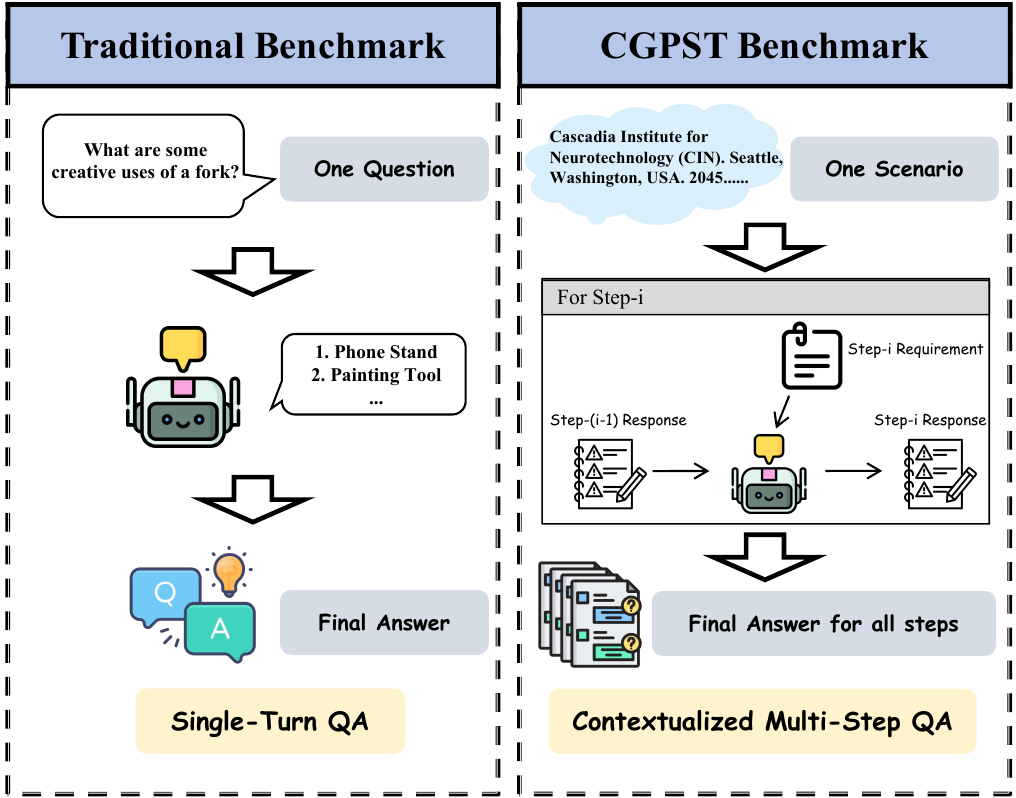}  
  \caption{Benchmark Comparison.}
  \label{fig:benchmark}
\end{figure}
Recently, researchers have begun exploring multi-Large Language Model (LLM) frameworks for solving contextualized tasks \citep{lu2024llmdiscussionenhancingcreativity, feng2025doctoragentrlmultiagentcollaborativereinforcement, liang-etal-2024-encouraging, lin2025creativityllmbasedmultiagentsystems}. For example, LLM Discussion enhances performance on creativity tasks such as Alternative Uses Tasks (AUT) and Instances Test by assigning different roles to multi-LLM \citep{lu2024llmdiscussionenhancingcreativity}. However, these frameworks do not explicitly emulate human team role division, which may lead to a single perspective and exacerbate output homogenization. This limitation may weaken performance on multi-step contextualized tasks \citep{Xu2025, wenger2025weredifferentweresame, lu2024llmdiscussionenhancingcreativity, 10.1145/3715928.3737479}. 
Moreover, frameworks such as LLM Discussion  \citep{lu2024llmdiscussionenhancingcreativity} are designed for single-step tasks and may not be directly applicable to multi-step contextualized tasks.

Multi-step contextualized tasks require not only diverse perspectives but also continuous interaction among LLMs to ensure \textbf{process controllability}, i.e., strict step-by-step adherence to task requirements. To this end, we propose \textbf{TeamLLM}, a human-like Team-Oriented Multi-LLM Collaboration Framework, introducing two innovations.

\textbf{(a) Team role division for diverse perspectives}. 
Inspired by the well-established sociological theory of Belbin’s team roles \citep{Belbin2010, Belbin2022}, we build TeamLLM based on this human team role theory for the first time and assign each LLM a distinct role with specific responsibilities, thereby enabling team role-driven collaboration. \textbf{(b) Three-phase collaboration for process controllability}. Building on team role-based division, we introduce a three-phase process: task initiation, perspective sharing, and consensus building. All LLMs participate in each task step, engaging in multi-turn interactions according to their roles within each step. This design ensures active communication and process.

To assess the effectiveness of TeamLLM in multi-step contextualized tasks, inspired by the Future Problem Solving Program \citep{treffinger2012four}, we propose \textbf{C}ontextually-\textbf{G}rounded and \textbf{P}rocedurally-\textbf{S}tructured \textbf{T}asks (\textbf{CGPST}) and build the benchmark. Existing contextualized task benchmarks, such as NOVELTYBENCH \citep{zhang2025noveltybench}, typically adopt a \textbf{single-step} format, requiring a one-shot answer without intermediate steps \citep{fang2025creationmmbenchassessingcontextawarecreative, lu-etal-2025-benchmarking, 10.1145/3613904.3642731, Ye_Gu_Zhao_Yin_Wang_2025}. Such settings make it difficult to evaluate LLMs comprehensively especially when tasks involve multi-step and process-oriented evaluation. 

As shown in Figure~\ref{fig:benchmark}, CGPST exhibits four core features. \textbf{Contextual Grounding}: Each task is set in a complete future scenario and require LLMs to consider multiple aspects rather than isolated problems. \textbf{Procedural Structure}: Tasks follow a coherent sequence, with each step depending on previous answers, enforcing logical consistency across steps. \textbf{Process-Oriented Evaluation}: Intermediate steps allow fine-grained assessment beyond final results, providing more comprehensive insights. \textbf{Multi-Dimensional Assessment}: Each step is scored across multiple dimensions within each step, capturing diverse abilities.

By combining TeamLLM with CGPST benchmark, we propose a new paradigm for systematically evaluate LLMs on multi-step contextualized tasks. We evaluate ten LLMs on CGPST, analyzing performance at overall, step and dimension levels. 

The contributions of this study are as follows.
\begin{itemize}[topsep=0pt, partopsep=0pt, parsep=0pt, itemsep=2pt]
  \item \textbf{TeamLLM}, a human-like multi-LLM collaboration framework for multi-step contextualized tasks via team role assignment.
  \item \textbf{The CGPST benchmark}, multi-step contextualized tasks to evaluate the effectiveness of proposed TeamLLM, with distinct scenarios, full-process responses and the corresponding human scores of ten LLMs.
  \item \textbf{Comparative analysis}, a comprehensive assessment of various LLMs on CGPST, providing detailed insights into their performance on multi-step contextualized tasks.
\end{itemize}


\begin{figure*}
    \centering
    \includegraphics[width=\textwidth]{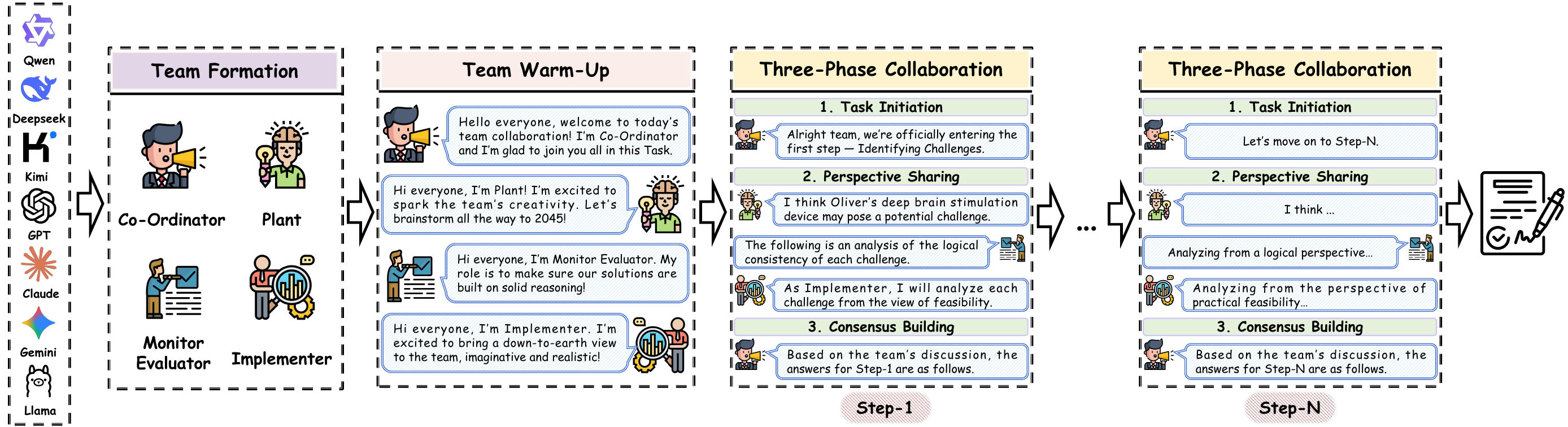} 
    \caption{TeamLLM: A Human-Like Team-Oriented Collaboration Framework.}
    \label{fig:TeamLLM}
\end{figure*} 

\section{Related Work}
\subsection{Multi-LLM Collaboration}

Rapid progress in LLMs has enabled various forms of collaboration design. Some approaches are function-oriented, assigning different modules to different LLMs \citep{gottweis2025aicoscientist, Wang2025, https://doi.org/10.48550/arxiv.2506.12508, bai2025twostepmultiagenttaskplanning}. For example, the three-phase travel planning framework improves performance with careful division and assignment  \citep{xie2024humanlikereasoningframeworkmultiphases}. Others explored interaction-oriented designs. The Multi-Agent Debate (MAD) framework, where LLMs debate in a "tit-for-tat" manner and a judge decides the outcome, improves logical accuracy \citep{10.5555/3692070.3692537, liang-etal-2024-encouraging}. LLM Discussion framework emphasizes discussion and idea integration, which promotes interaction and creativity through a three-phase discussion \citep{lu2024llmdiscussionenhancingcreativity}. Together, these approaches provide useful insights into multi-LLM coordination, yet they generally do not explicitly emulate human-like team roles, and are not specifically designed for multi-step contextualized tasks.

\subsection{Benchmarking LLMs on Contextualized Tasks}
\begin{table}[h]
    \scriptsize
	\begin{tabular}{l c c}\toprule
		\textit{}  & \textit{\scriptsize Multi-step}  & \textit{\scriptsize Evaluation} \\ \midrule
		\textbf{CGPST} & \ding{51} & process-oriented \\

        TTCW \citep{10.1145/3613904.3642731} & \ding{55} & result-oriented \\ NOVELTYBENCH \citep{zhang2025noveltybench} & \ding{55} & result-oriented \\ 
		CREATIVEMATH \citep{Ye_Gu_Zhao_Yin_Wang_2025} & \ding{55} & result-oriented \\
		AidanBench \citep{mclaughlin2024aidanbench} & \ding{55} & result-oriented \\
        LiveIdeaBench \citep{ruan2025liveideabenchevaluatingllmsdivergent} & \ding{55} & result-oriented \\ 
        Creation-MMBench \citep{fang2025creationmmbenchassessingcontextawarecreative} & \ding{55} & result-oriented \\
        NEOCODER \citep{lu-etal-2025-benchmarking} & \ding{55} & result-oriented \\
        \bottomrule
	\end{tabular}
    \caption{Comparison between the CGPST benchmark and several contextualized benchmarks.}
    \label{tab:benchmark}
\end{table}

Benchmarking is the standard method for evaluating LLMs. Various benchmarks target specific abilities: CREATIVEMATH \citep{Ye_Gu_Zhao_Yin_Wang_2025} for generating novel mathematical solutions and TTCW \citep{10.1145/3613904.3642731} for creative story generation from a plot. While valuable, they have three limitations. First, they rely on simple scenarios (e.g., a single math problem or a story plot). Second, their single-step format lacks the coherence across steps, so correct answers may mask underlying flaws. Third, without a multi-step design, evaluations focus solely on final outputs, hindering fine-grained analysis of intermediate process and comprehensive assessment of LLMs' abilities.


\section{TeamLLM Framework}
 Inspired by Belbin’s theory of team roles \citep{Belbin2010, Belbin2022}, we adopt four team roles: \textbf{Co-Ordinator}, \textbf{Plant}, \textbf{Monitor Evaluator} and \textbf{Implementer} (CO, PL, ME, IMP) to form a multi-LLM team, denoted as \textbf{MLTeam}. As shown in Figure~\ref{fig:TeamLLM}, MLTeam first undergo a \textbf{Warm-Up} phase for initial activation. Then, for each CGPST step, MLTeam operates under a \textbf{three-phase collaboration framework}. This multi-turn interaction can enhance inter-LLM collaboration, promote diverse ideas, and ensure process controllability.
 
\subsection{Team Formation}
In this phase, each LLM is assigned a distinct team role via a triplet \{\textit{Team\_Role}, \textit{Role\_Speciality}, \textit{Role\_Prompt}\}, which defines its speciality and behavioral guidelines \citep{lu2024llmdiscussionenhancingcreativity}, as detailed in Appendix~\ref{app:role-play}. This role-play mechanism enables LLMs to propose, evaluate, and coordinate solutions across the collaboration, emulating human-like division of interaction.

\begin{algorithm}[t]
{\fontsize{8.4pt}{12pt}\selectfont
\caption{Pseudo code of Team Warm-Up}
\label{alg:warmup}
\begin{algorithmic}
\State \textbf{Input:} Role-Play prompt for each LLM, warm\_up prompt
\State Initialize step\_history = [] \textcolor{gray}{\scriptsize \# Store discussion history of current step.}
\State $O_{CO}^{(0)} = A_{CO}.\text{ask}(\text{Role\_Play prompt + warm\_up prompt})$
\State \text{Update step\_history}$\xleftarrow{\text{append}}O_{CO}^{(0)}$
\For{each $i \in \{PL, ME, IMP\}$} \textcolor{gray}{\scriptsize \# Each LLM speaks in turn.}
    \State $O_i^{(0)} = A_i.\text{ask}(\text{Role\_Play prompt} + \text{step\_history})$
    \State \text{Update step\_history} $\xleftarrow{\text{append}}O_{i}^{(0)}$
\EndFor
\end{algorithmic}
}
\end{algorithm}

\subsection{Team Warm-Up}
To help MLTeam members get to know each other and clarify their responsibilities, we introduce a Warm-Up phase after team formation as an initial activation \cite{lu2024llmdiscussionenhancingcreativity}. This phase facilitates early collaboration, establishes shared context, and lay the foundation for subsequent tasks.

Algorithm~\ref{alg:warmup} shows the Warm-Up procedure. Here, $O_{i}^{(s)}$ denotes the response of role $i$ at step $s$, with $s=0$ indicating the Warm-Up phase. $A_{i}$ denotes the LLM assigned to role $i$. $ask(p)$ and $Update\ step\_history(o)$ indicates the response generated from information $p$ and the operation that appends message $o$ to $step\_history$, respectively. The $warm\_up \ prompt$ guides the CO to initiate MLTeam Warm-Up, as shown in Appendix~\ref{app:TeamLLM Prompts Variables}.

\subsection{Three-Phase Collaboration}
To ensure process controllability, we design a three-phase collaboration framework, as shown in Algorithm~\ref{alg:three_phase}. The variable $log\_history$ stores aggregated team answers from previous steps. $step\_prompt$ provides task information for each step, where $N$ denotes the total number of steps. Each $SP_i$ is represented as a quadruple \{\textit{Step\_Number}, \textit{Step\_Name}, \textit{Step\_Description}, \textit{Step\_Output}\}, with full details in Appendix~\ref{app:CGPST step prompt}.

\subsubsection{Task Initiation}
In this phase, the CO receives the task information for the current step and conveys it to MLTeam, clarifying objectives and requirements and guiding the team into a discussion state.
\begin{equation}
\scalebox{0.9}{$
\begin{split}
O_{CO}^{(s)} = A_{CO}.ask(
Role\_Play\ prompt + \\
Task\_Initiation\ prompt
)
\end{split}
$}
\end{equation}
\begin{equation}
\scalebox{0.9}{$
step\_history = Update(O_{CO}^{(s)})
$}
\end{equation}

\begin{algorithm}[t]
{\fontsize{8pt}{12pt}\selectfont
\caption{{\fontsize{9.5pt}{9pt}\selectfont Pseudo code of Three-Phase Collaboration}}
\label{alg:three_phase}
\begin{algorithmic}
\State \textbf{Input:} Role-Play prompt for each LLM, step prompt = (SP$_1$, SP$_2$, ..., SP$_N$), Task\_Initiation prompt, Consensus\_Building prompt
\State Initialize log\_history = [] \textcolor{gray}{\scriptsize \# Store answers of each step.}
\For{each $s \in \{1,2,...,N\}$} \textcolor{gray}{\scriptsize \# MLTeam perform each step in order.}
    \State Initialize step\_history = [] \textcolor{gray}{\scriptsize \# Store discussion history of current step.}
    \State \textcolor{gray}{\scriptsize \# Three-Phase Collaboration}
    \State \text{Update step\_history}$\xleftarrow{\text{append}}$
    \State \hspace{1.5em} \text{Task\_Initiation}($A_{CO}$,SP$_s$,\text{Task\_Initiation prompt})
    \State \text{Update step\_history}$\xleftarrow{\text{append}}$
    \State \hspace{1.5em} \text{Perspective\_Sharing}($A_{PL},A_{ME},A_{IMP}$)
    \State \text{Update log\_history}$\xleftarrow{\text{append}}$
    \State \hspace{1.5em} \text{Consensus\_Building}($A_{CO}$,\text{Consensus\_Building prompt})
\EndFor
\end{algorithmic}
}
\end{algorithm}

\begin{table*}
    \centering
    \scriptsize
    \begin{tabular}{>{\footnotesize}p{2.2cm}>{\footnotesize}p{6.5cm}>{\footnotesize}p{5cm}>{\footnotesize}p{0.6cm}}
        \toprule
        \textbf{Step} & \textbf{Requirement} & \textbf{Evaluation Metric} & \textbf{Score} \\
        \midrule
        Step-1: Identify Challenges & Identify up to 8 reasonable challenges based on the future scenario. & Fluency, Flexibility, Elaboration, Originality & 48 \\
        Step-2: Select an Underlying Problem & Select the most promising and meaningful challenge from Step-1 as the underlying problem. & Integrity of Underlying Problem, Focus of Underlying Problem, Importance / Adequacy of Underlying Problem & 30 \\
        Step-3: Produce Solutions & Generate up to 8 solutions for the underlying problem from Step-2. & Fluency, Flexibility, Elaboration, Originality & 48 \\
        Step-4: Select Criteria & Generate 5 evaluation criteria for the solutions from Step-3. & Correctly Written, Relevance to the Underlying Problem & 20 \\
        Step-5: Apply Criteria to Top Solution & Rank the solutions from Step-3 using the criteria from Step-4 and select the highest-scoring solution. & Correctly Used & 5 \\
        Step-6: Develop an Action Plan & Develop the top solution from Step-5 into a comprehensive action plan to address the underlying problem from Step-2. & Relevance, Effectiveness, Criteria in Development of Action Plan, Impact, Humaneness, Development of Action Plan & 35 \\
        \bottomrule
    \end{tabular}
    \caption{Detailed Step-wise Information and Evaluation Metrics on CGPST}
    \label{tab:CGPST}
\end{table*}

\subsubsection{Perspective Sharing}
First, the PL generates an initial solution based on the task requirements, providing a preliminary basis for team discussion. The ME then reviews and refines it to ensure logical consistency, followed by the IMP evaluating its practical feasibility. Through this multi-perspective refinement, MLTeam gradually builds a more complete solution.

\scalebox{0.9}{$\forall i \in \{PL, ME, IMP\}:$}
\begin{equation}
\scalebox{0.9}{$
\begin{split}
O_i^{(s)} = A_i.ask(Role\_Play\ prompt \\+ log\_history + step\_history)
\end{split}
$}
\end{equation}
\begin{equation}
\scalebox{0.9}{$
step\_history = Update(O_i^{(s)})
$}
\end{equation}

\subsubsection{Consensus Building}
The CO integrates perspectives from the previous phase to produce the final team answer of the current step and the result $Answer^{(s)}$ is appended to $log\_history$ as part of historical record, providing continuous context for future steps. This ensures MLTeam's output is coherent and holistic, providing a solid foundation for subsequent collaboration. 
\begin{equation}
\scalebox{0.9}{$
\begin{split}
Answer^{(s)} = A_{CO}.ask(Role\__Play\ prompt\\ + log\_history\ + step\_history\\ + Consensus\_Building\ prompt)
\end{split}
$}
\end{equation}
\begin{equation}
\scalebox{0.9}{$
log\_history=Update(Answer^{(s)})
$}
\end{equation}

on CGPST, ensuring cross-step coherence and multi-turn collaboration within each step imposes specific demands on context management. Although modern LLMs support large context windows (up to 64K or 1M tokens), simply concatenating all historical records can cause instruction forgetting or underutilization, hindering effective task progression \citep{gao-etal-2024-insights, lu2024controlledstudylongcontext}. 

To address this, TeamLLM uses a \textbf{dual-variable context management mechanism}. Across steps, $log\_history$ stores the final answer of each step to ensure \textbf{cross-step coherence}. Within a step, $step\_history$ preserves the ongoing discussion, supporting \textbf{within-step collaboration}. This reduces unnecessary context accumulation while retaining critical information, maintaining both contextual integrity and process controllability.


\section{CGPST Benchmark}

\begin{table*}[htbp]
    \centering
    {\fontsize{7}{10}\selectfont  
    \begin{tabular}{p{1.7cm} p{13.3cm}}  
        \toprule
        \textbf{Category} & \textbf{Model (Abbreviation)} \\
        \midrule
        Open-Source & qwen3-235b-a22b-instruct-2507 (\texttt{qwen3-instruct}), qwen3-235b-a22b-thinking-2507 (\texttt{qwen3-thinking}), DeepSeek-V3-0324 (\texttt{deepseek-v3}), Deepseek-r1 (\texttt{deepseek-r1}), kimi-k2-0711-preview (\texttt{kimi-k2}), llama-4-scout-17b-16e-instruct (\texttt{llama-4-scout}) \\
        \midrule
        Closed-Source & gpt-4o (\texttt{gpt-4o}), gpt-5 (\texttt{gpt-5}), claude-opus-4-1-20250805 (\texttt{claude-4-opus}), gemini-2.5-pro (\texttt{gemini-2.5-pro}) \\
        \bottomrule
    \end{tabular}
    }
    \caption{LLMs Selected for Evaluation}
    \label{tab:all_models}
\end{table*}

\subsection{Contextually Grounding and Procedurally Structuring} 
Each task is set within a future scenario, requiring models to generate answers consistent with contextual information. The \textbf{10 scenarios} (FS1-FS10) cover topics such as biosecurity, providing a broad and practically relevant set of challenges.

Inspired by the Future Problem Solving Program \cite{treffinger2012four} and the Creative Problem Solving (CPS) model \cite{Treffinger1995}, each task consists of \textbf{six sequential steps} as shown in Table~\ref{tab:CGPST}, where the output of one step will influence the next. This structure allows a comprehensive assessment of multi-step thinking and problem solving.

\subsection{Multidimensional Ability Assessment} 
\label{Section 4.2}
The benchmark evaluates both final answers and step-wise performance, enabling process-oriented assessment and nuanced comparison of distinct LLMs. Detailed dimensions are in Appendix~\ref{app:rubrics}.

\noindent\textbf{Step-1 and Step-3} 

These steps require generating multiple challenges or solutions, thereby assessing \textbf{divergent thinking}. We adopt the four classical dimensions from the Torrance Tests of Creative Thinking (TTCT) \cite{torrance1966torrance}, widely used in human creativity assessment \cite{10.1145/3706598.3714198, Zhao2025}. Scores are calculated for each item. Given LLMs’ tendency to produce large quantities of responses, \textit{Fluency} and \textit{Flexibility} are less discriminative, while \textit{Elaboration} and \textit{Originality} better capture creative quality \cite{lu2024llmdiscussionenhancingcreativity}; thus, a 3-point scale is used for these two dimensions. More details are shown in Appendix~\ref{app:dimension details for step-1 and Step-3}.
\begin{itemize}[noitemsep, topsep=0pt]
  \item \textbf{Fluency}: number of reasonable responses.
  \item \textbf{Flexibility}: diversity of categories covered.
  \item \textbf{Elaboration}: degree of detail in descriptions.
  \item \textbf{Originality}: novelty of responses.
\end{itemize}

\noindent\textbf{Step-2} 

Step-2 focuses refining a single underlying problem (UP), reflecting \textbf{convergent thinking}.
\begin{itemize}[noitemsep, topsep=0pt]
  \item \textbf{Integrity}: completeness of UP description.
  \item \textbf{Focus}: logical rigor and consistency of UP.
  \item \textbf{Importance}: significance of UP.
\end{itemize}

\noindent\textbf{Step-4 and Step-5}

These involve designing and applying criteria, assessing \textbf{critical thinking and logical thinking}.
\begin{itemize}[noitemsep, topsep=0pt]
  \item \textbf{Correctly Written}: valid criteria.
  \item \textbf{Relevance}: alignment with the UP.
  \item \textbf{Correctly Used}: correct application of criteria to rank all solutions without ties.
\end{itemize}

\noindent\textbf{Step-6}

This step develops the optimal solution from Step-5 into a comprehensive action plan, evaluating \textbf{comprehensive problem solving}.
\begin{itemize}[noitemsep, topsep=0pt]
  \item \textbf{Relevance}: alignment with the UP.
  \item \textbf{Effectiveness}: ability to address the UP.
  \item \textbf{Criteria Alignment}: adherence to criteria.
  \item \textbf{Impact}: positive effect on the scenario.
  \item \textbf{Humaneness}: attention to human factors.
  \item \textbf{Development}: completeness and feasibility.
\end{itemize}

A complete CGPST scenario (FS10), the six-step responses of \texttt{kimi-k2}, and the corresponding human scores are provided in Appendix~\ref{app:complete CGPST example}.


\begin{table*}[htbp]
\centering
{\fontsize{8pt}{10pt}\selectfont
\begin{tabular}{l*{10}{c}c}
\toprule
\textbf{Model} & \textbf{FS1} & \textbf{FS2} & \textbf{FS3} & \textbf{FS4} & \textbf{FS5} & \textbf{FS6} & \textbf{FS7} & \textbf{FS8} & \textbf{FS9} & \textbf{FS10} & \textbf{Avg} \\
\midrule
qwen3-instruct - TeamLLM & \textbf{135.5} & \textbf{142} & \textbf{127} & \textbf{123.5} & \textbf{120.5} & \textbf{142} & \textbf{156} & \textbf{135.5} & \textbf{137} & \textbf{134.5} & \textbf{135.35}$^{**}$ \\
qwen3-instruct - baseline & 116 & 113 & 110 & 122.5 & 107.5 & 127.5 & 125.5 & 113 & 117 & 110 & 116.2 \\
\cdashline{1-12}
qwen3-thinking - TeamLLM & \textbf{115} & \textbf{133.5} & \textbf{135} & \textbf{124} & \textbf{117.5} & \textbf{140.5} & \textbf{146.5} & 111 & 116 & 112 & \textbf{125.1}$^{**}$ \\
qwen3-thinking - baseline & 100.5 & 108.5 & 121 & 120.5 & 115 & 126.5 & 130.5 & \textbf{111.5} & 103.5 & 108 & 114.55 \\
\cdashline{1-12}
deepseek-v3 - TeamLLM & \textbf{129} & \textbf{137} & \textbf{130.5} & \textbf{116.5} & \textbf{124.5} & \textbf{114} & \textbf{121.5} & 102.5 & \textbf{124.5} & \textbf{130} & \textbf{123}$^{**}$ \\
deepseek-v3 - baseline & 114.5 & 104.5 & 91 & 104.5 & 110 & 110 & 116 & \textbf{105.5} & 104 & 109.5 & 106.95 \\
\cdashline{1-12}
deepseek-r1 - TeamLLM & \textbf{126.5} & \textbf{132.5} & 114.5 & \textbf{131} & \textbf{135.5} & \textbf{120.5} & \textbf{126} & \textbf{116} & \textbf{119} & 107 & \textbf{122.85}$^{**}$ \\
deepseek-r1 - baseline & 109.5 & 99 & 109.5 & 104 & 97.5 & 110 & 119 & 106.5 & 103 & \textbf{112} & 107 \\
\cdashline{1-12}
kimi-k2 - TeamLLM & \textbf{134} & \textbf{134} & \textbf{133} & 120 & \textbf{122} & \textbf{129} & 107.5 & 117 & \textbf{127} & \textbf{116.5} & \textbf{124} \\
kimi-k2 - baseline & 108 & 113.5 & 119 & \textbf{124.5} & 110 & 127.5 & \textbf{129} & \textbf{124.5} & 116.5 & 115 & 118.75 \\
\cdashline{1-12}
llama-4-scout - TeamLLM & 101 & \textbf{120.5} & \textbf{129} & 91 & \textbf{111} & \textbf{116} & 105.5 & 93 & \textbf{107.5} & \textbf{120} & \textbf{109.45} \\
llama-4-scout - baseline & \textbf{101.5} & 86 & 105.5 & \textbf{105.5} & 91.5 & 101 & \textbf{110} & \textbf{106} & 71 & 104.5 & 98.25 \\
\cdashline{1-12}
gpt-4o - TeamLLM & \textbf{137} & \textbf{134.5} & \textbf{118} & \textbf{121} & \textbf{136.5} & \textbf{138.5} & 124 & \textbf{135} & \textbf{118} & \textbf{123} & \textbf{128.55}$^{**}$ \\
gpt-4o - baseline & 102 & 91 & 104.5 & 109.5 & 96 & 118 & 124 & 109 & 94.5 & 110 & 105.85 \\
\cdashline{1-12}
gpt-5 - TeamLLM & \textbf{144} & \textbf{139.5} & \textbf{139.5} & \textbf{138} & \textbf{156} & \textbf{143} & \textbf{156} & \textbf{150} & \textbf{130} & \textbf{132} & \textbf{142.8}$^{**}$ \\
gpt-5 - baseline & 133 & 126 & 130.5 & 127 & 130 & 126.5 & 142 & 139 & 119 & 127.5 & 130.05 \\
\cdashline{1-12}
claude-4-opus - TeamLLM & \textbf{133.5} & \textbf{134.5} & \textbf{140.5} & \textbf{132} & \textbf{138.5} & \textbf{144.5} & \textbf{150.5} & \textbf{147.5} & \textbf{116.5} & \textbf{129} & \textbf{136.7}$^{**}$ \\
claude-4-opus - baseline & 108.5 & 100 & 123.5 & 128 & 118.5 & 127.5 & 133 & 123.5 & 109 & 114.5 & 118.6 \\
\cdashline{1-12}
gemini-2.5-pro - TeamLLM & \textbf{130.5} & \textbf{123} & 121.5 & 117.5 & \textbf{143} & \textbf{142.5} & 117.5 & \textbf{143} & 116 & \textbf{120} & \textbf{127.45} \\
gemini-2.5-pro - baseline & 115.5 & 111 & \textbf{124.5} & \textbf{124.5} & 115 & 121.5 & \textbf{123.5} & 118 & 116 & 118 & 118.75 \\
\bottomrule
\end{tabular}
\caption{Overall performance across 10 CGPST scenarios with Wilcoxon signed-rank test results. Bold indicates the higher score. “Avg” is the model’s mean score, and $^{**}$ marks a significant improvement under TeamLLM (p=0.01).}
\label{tab:overall_performance}
}
\end{table*}

\section{Experiments}
\subsection{Experiment Setup}
We set single-LLM as the baseline to evaluate TeamLLM's effectiveness using CGPST across ten LLMs (Table~\ref{tab:all_models}), while comparing different LLMs' performance. These models vary in source type and optimization focus, ensuring broad representativeness. More experimental details are shown in Appendix~\ref{app:hyperparameter}. Additionally, during Team Formation, meta prompts convey key information, including the scenario and task requirements (Appendix~\ref{app:meta prompt}).

\subsection{Human Evaluation}
\label{section 5.3}
We recruit 15 trained human experts to evaluate all responses. Each response is scored by two experts in a double-blind setup. Inter-rater consistency is quantified with: (a) \textbf{Pearson correlation coefficient (PCC)}, measuring linear agreement between two raters on the same response and (b) \textbf{Intra-class correlation coefficient (ICC)}, evaluating absolute agreement across all responses in a scenario. A calibration mechanism ensure reliability: for responses with PCC below 0.7 (practically 0.65), a third expert reassesses and final scores are adjusted. Table~\ref{tab:human_evaluation_consistency} shows the improvements, demonstrating the effectiveness of third-party calibration. Full details are shown in Appendix~\ref{app:human evaluation}.
\begin{table}[t]
\centering
\scriptsize
\begin{tabular}{l c c}
\toprule
\textbf{Model} & \textbf{TeamLLM} & \textbf{Baseline} \\
\midrule
qwen3-thinking & 1.7 & 0   \\
kimi-k2             & 2.4 & 0   \\
gemini-2.5-pro      & 2.9 & 0.4 \\
\bottomrule
\end{tabular}
\caption{Average number of blank responses in Step-3.}
\label{tab:step3_blank}
\end{table}
\begin{table}[htbp]
\centering
\scriptsize
\begin{tabular}{lcc}
\toprule
\textbf{Metric} & \textbf{Before Calibration} & \textbf{After Calibration} \\
\midrule
PCC & 0.71 & 0.87 \\
ICC & 0.58 & 0.84 \\
\bottomrule
\end{tabular}
\caption{Consistency Before and After Calibration.}
\label{tab:human_evaluation_consistency}
\end{table}


\section{Results}
\subsection{Overall Performance}
Table~\ref{tab:overall_performance} shows TeamLLM consistently improves performance across all scenarios, with seven models showing statistically significant gains, showing the potential of team-oriented approach.

Improvement varies by model. Models with high performance (\texttt{claude-4-opus}, \texttt{qwen3-instruct}, \texttt{GPT-5}) gain the most, while \texttt{llama-4-scout} and \texttt{kimi-k2} improve moderately, suggesting intrinsic model ability influences team effectiveness. Within the \texttt{Qwen} series, the \texttt{instruct} outperforms the \texttt{thinking}, likely due to stronger instruction-following training that better aligns with structured multi-step tasks requirements. 

\subsection{Step-level Comparison}
\begin{figure*}
    \centering
    \includegraphics[width=0.9\textwidth]{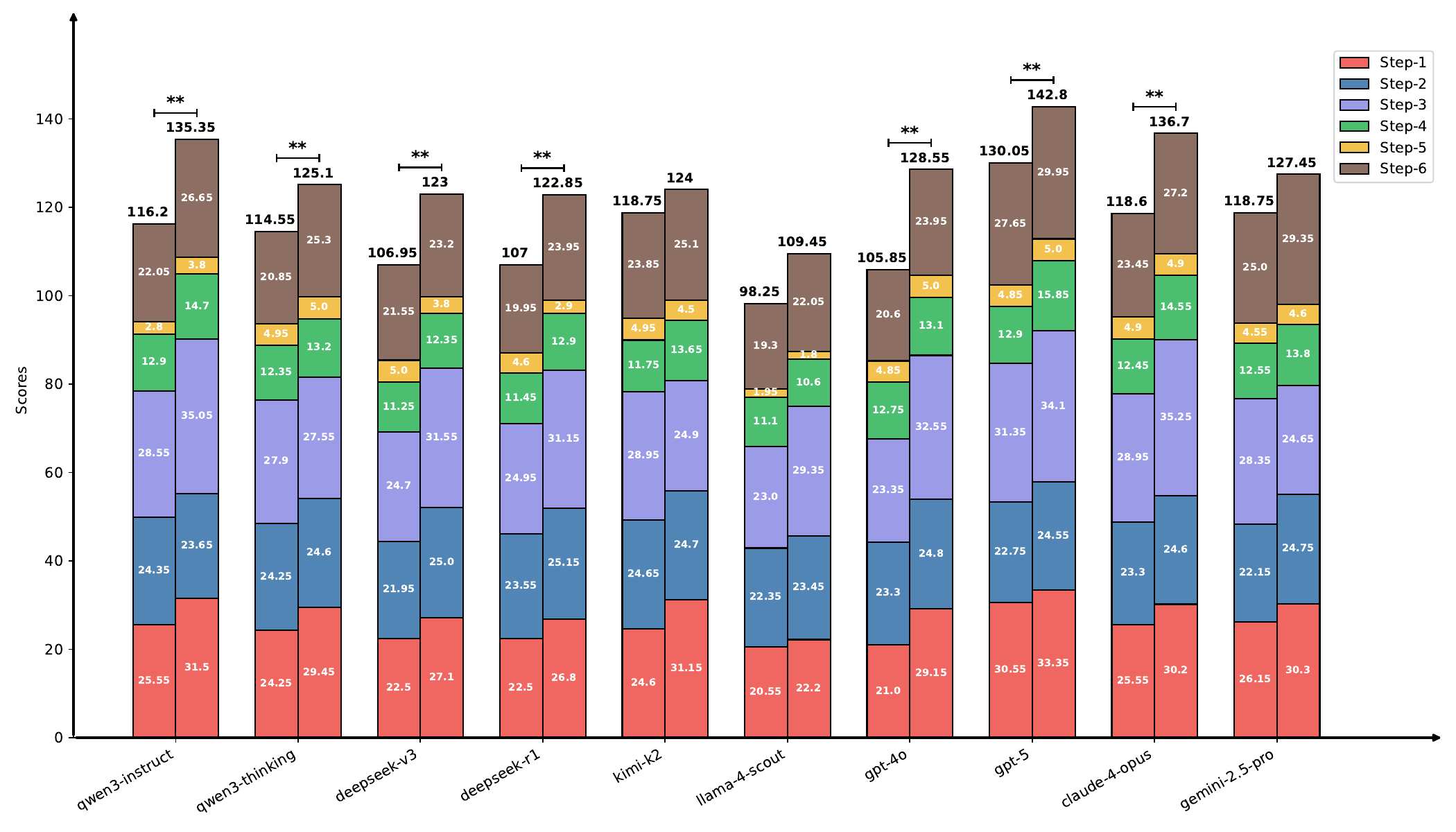} 
    \caption{Step-level performance of TeamLLM (right bars) and the baseline (left bars), with different colors representing distinct steps. Horizontal lines with $^{**}$ indicate statistically significant differences at p=0.01.}
    \label{fig:step-level}
\end{figure*}

To examine how TeamLLM improves performance, we conduct a step-level analysis. Figure~\ref{fig:step-level} shows that TeamLLM generally boosts performance across all steps, with the largest gains in divergent thinking (Step-1, 3) and action plan development (Step-6). Improvements in Step-2 are smaller, reflecting relatively strong convergent abilities even under the baseline. Notably, \texttt{kimi-k2} and \texttt{gemini-2.5-pro} show slight declines in Step-3, which will be analyzed in Section~\ref{section 7.1}.

Examining step-wise differences highlights model-specific strengths. In Step-1 and 3, \texttt{gpt-5}, \texttt{qwen3-instruct}, and \texttt{claude-4-opus} excel at producing diverse challenges and feasible solutions. Step-2 shows minimal variation, reflecting balanced problem-focusing abilities. Steps-4 and 5 exhibit larger score differences, which will be analyzed in Section~\ref{section 6.3.3}. Step-6 evaluates action plan formulation, where \texttt{gpt-5} achieves the highest scores under both conditions, reinforcing its comprehensive ability on CGPST.

\subsection{Dimension-level Analysis within Steps}
We analyze model performance across individual dimensions for each step, as shown in appendix~\ref{app:dimension-level rader}.

\subsubsection{Step-1 and Step-3: Divergent Thinking}
In divergent thinking steps, TeamLLM primarily enhances \textit{Elaboration} and \textit{Originality}. For example, \texttt{qwen3-instruct} improves in Step-3 from 12.05 to 14.35 (\textit{Elaboration}) and 4.65 to 8.65 (\textit{Originality}), while \texttt{gpt-4o} rises in Step-1 from 6.65 to 10.05 and 1.85 to 5.5, respectively. \textit{Fluency} and \textit{Flexibility} are already strong under the baseline, so the main gains are on \textbf{\textit{Elaboration}} and \textbf{\textit{Originality}}, which are more indicative of divergent thinking \cite{lu2024llmdiscussionenhancingcreativity}, highlighting the effectiveness of the framework in promoting creative output.

\subsubsection{Step-2: Convergent Thinking}
For convergent thinking, overall improvements are modest, as most models already perform well under the baseline. TeamLLM’s main advantage is in the \textit{Focus} dimension, where team collaboration enables models to better identify and prioritize critical information within the scenario, thereby enhancing problem recognition and foresight.

\subsubsection{Step-4 and Step-5: Critical Thinking and Logical Thinking}
\label{section 6.3.3}
In Step-4, improvements mainly appear in \textit{Relevance}, indicating that TeamLLM helps align evaluation criteria more closely with the scenario rather than relying on generic standards, enhancing contextual appropriateness. In Step-5,  TeamLLM generally stabilizes performance and benefits some LLMs, but not all: \texttt{deepseek-v3} and \texttt{deepseek-r1} show declines, suggesting that collaboration may introduce reasoning conflicts for models with weaker logical chains, whereas strong models gain greater consistency and accuracy.

Cross-model comparison reveals performance tiers. \texttt{claude-4-opus}, \texttt{qwen3-thinking}, \texttt{gpt-4o}, and \texttt{gpt-5} score near-perfect, indicating strong logical abilities. \texttt{llama-4-scout} ranks lowest (1.8–1.95), reflecting difficulty with criteria ranking. Duplicates and omissions cause its poor performance, with examples in Appendix~\ref{app:step-5 instances of llama}.

\begin{figure*}
    \centering
    \includegraphics[width=0.9\textwidth]{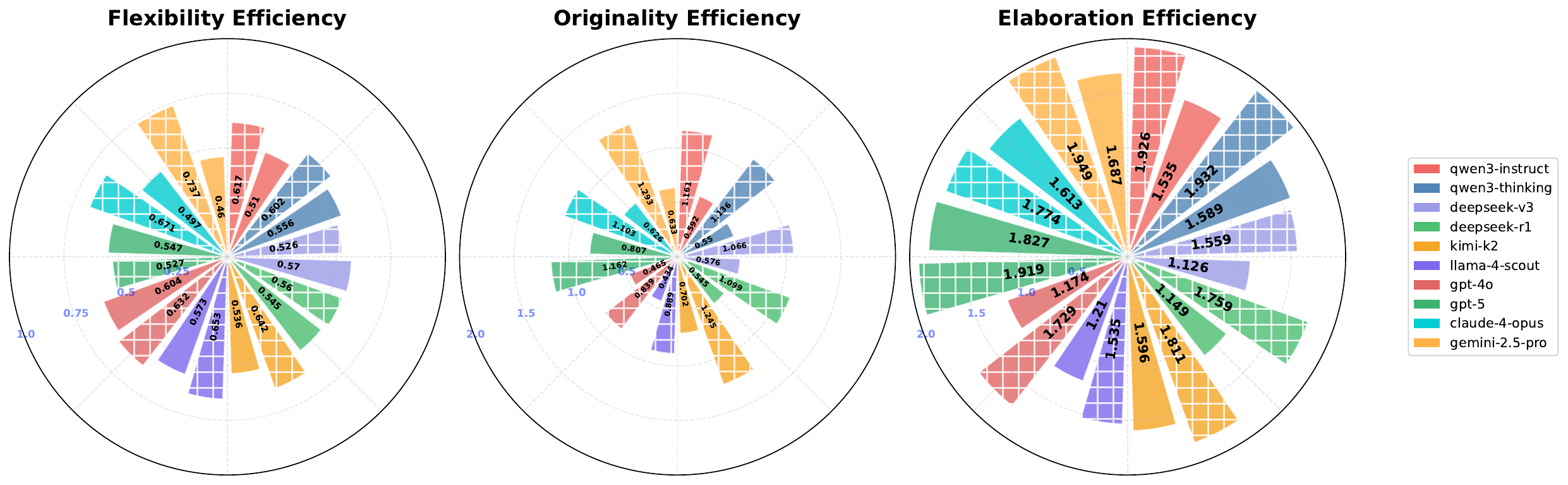} 
    \caption{Comparison of model performance in Step-3 across Flexibility Efficiency, Originality Efficiency, and Elaboration Efficiency. Solid bars correspond to the baseline, while hatched bars correspond to TeamLLM.}
    \label{fig:average Step-3}
\end{figure*} 

\subsubsection{Step-6: Comprehensive Problem Solving}
Step-6 requires developing a coherent action plan, aligned with earlier steps and effectively addressing the underlying problem in Step-2. This tests LLMs’ consistency and problem solving ability in multi-step tasks. TeamLLM generally improves performance across all dimensions, largely due to the diverse perspectives from team roles. \texttt{gpt-5} remains the top performer, demonstrating strong multi-step thinking, whereas \texttt{kimi-k2} shows limited gains, as longer collaborative contexts under TeamLLM constrain its improvement.

\begin{figure}[t]  
  \centering
  \includegraphics[width=\columnwidth]{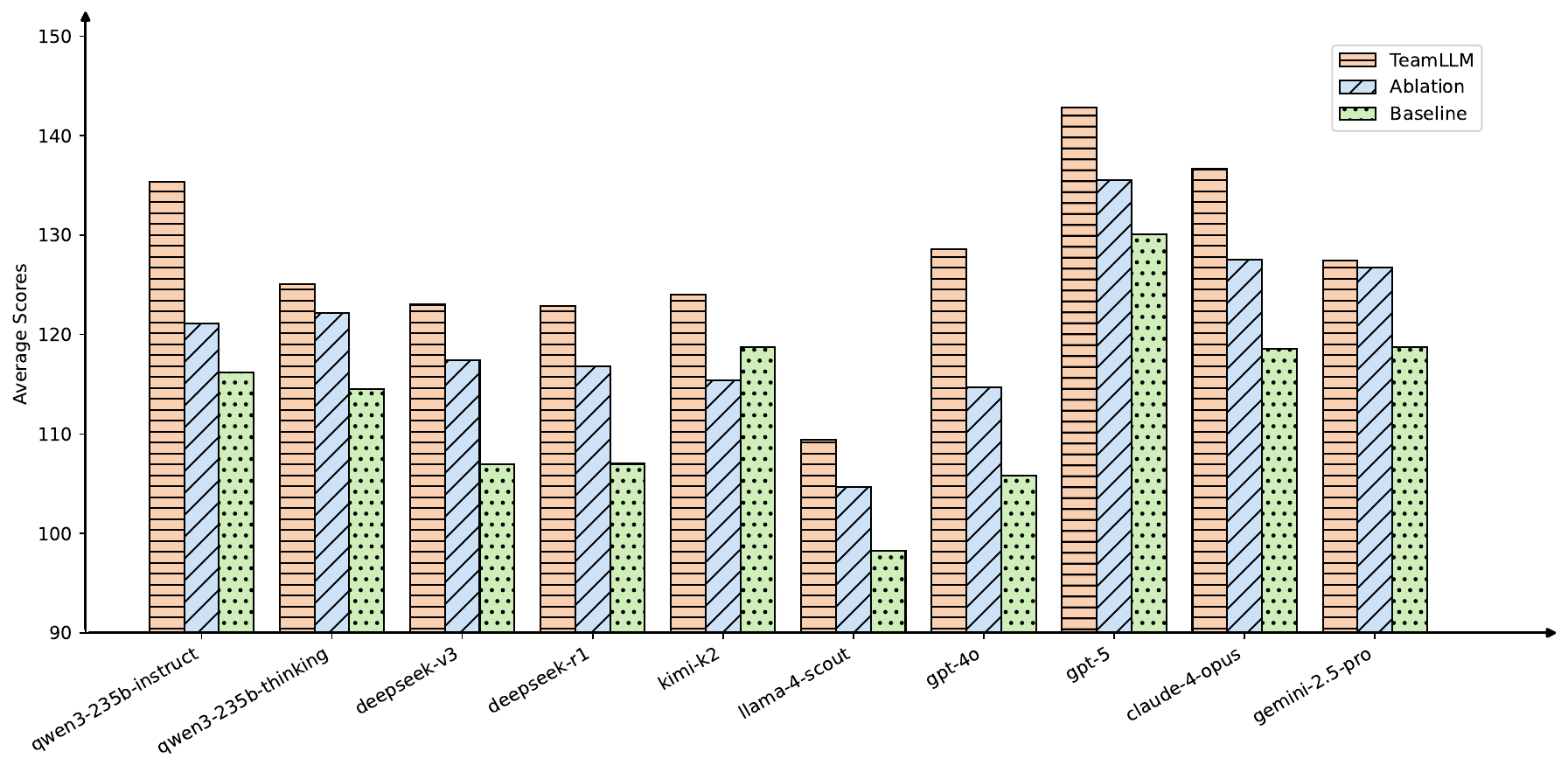}  
  \caption{Ablation study results.}
  \label{fig:ablation_results}
\end{figure}

\subsection{Ablation Study}
We conduct an ablation study to isolate the effect of team role-based collaboration in TeamLLM by removing explicit role assignments while keeping the number of LLMs unchanged, as shown in Figure~\ref{fig:ablation_results}. Removing explicit team role division leads to a consistent drop in performance, while still outperforming the baseline. The results confirm that the introduction of diverse perspectives derived from human team roles can enhances the performance of TeamLLM in contextualized multi-step tasks.

\section{Discussion}

\subsection{Divergent Thinking Quality and Efficiency Analysis in TeamLLM}
\label{section 7.1}
\texttt{qwen3-thinking}, \texttt{gemini-2.5-pro} and \texttt{kimi-k2} decline in Step-3 (Figure~\ref{fig:dimension-level}(c)), mainly due to each LLM producing 2–3 blank responses on average (Table~\ref{tab:step3_blank}). To account for this, we calculate per-solution efficiencies for Step-3 dimensions:
\begin{equation}
\scalebox{0.9}{$
Flexibility\_Efficiency=\frac{Flexibility}{Fluency}\in[0,1]
$}
\end{equation}
\begin{equation}
\scalebox{0.9}{$
Elaboration\_Efficiency=\frac{Elaboration}{Fluency}\in[0,2]
$}
\end{equation}
\begin{equation}
\scalebox{0.9}{$
Orginality\_Efficiency=\frac{Orginality}{Fluency}\in[0,2]
$}
\end{equation}

These metrics normalize for solution quantity, providing a clearer measure of solution quality. Figure~\ref{fig:average Step-3} shows that TeamLLM consistently improves \textit{Elaboration Efficiency} and \textit{Originality Efficiency}, confirming that collaborative team roles enhance each solution's creativity. Gains in \textit{originality} slightly exceed those in \textit{elaboration}, reflecting that LLMs can produce detailed ideas but still face limitations in novelty \cite{Zhao2025}.

\texttt{qwen3-thinking}, \texttt{gemini-2.5-pro} and \texttt{kimi-k2} achieve relatively high per-solution quality despite a drop in overall Step-3 scores, revealing a "\textbf{quantity–quality trade-off}": fewer generated solutions lead to lower absolute scores, but the quality of each solution improve. Notably, \texttt{gemini-2.5-pro} achieves the highest per-solution scores across all three dimensions, highlighting its strong divergent thinking ability.

\subsection{Mitigating Model Output Homogenization via TeamLLM: Insights from Step-1 and 3}
This section examines Step-1 and 3, both divergent thinking, to assess how TeamLLM reduces output homogenization. We quantify semantic diversity using \textbf{Self-BLEU} \cite{10.1145/3209978.3210080, Yin2020, lu2024llmdiscussionenhancingcreativity, liang-etal-2024-encouraging}, where higher Self-BLEU means lower diversity. Due to the negligible probability of 3-grams and 4-grams in Chinese, we compute Self-BLEU using a weighted 1-gram (0.8) and 2-gram (0.2) strategy. Diversity is defined as follows, where $\overline{Self\_BLEU(M)}$ is the average similarity of model $M$'s responses, and $C$ is the set of $N$ responses.
\begin{equation}
\scalebox{0.9}{$
\begin{split}
Diversity &= 1 - \overline{Self\_BLEU(M)} \nonumber \\
&=1-{\tfrac{1}{N}}\sum_{i=1}^N Self\_BLEU(c_i)
\end{split}
$}
\end{equation}
\begin{equation}
\scalebox{0.9}{$
Self\_BLEU(c_i)=BLEU(c_i,{c_j | j\neq i})
$}
\end{equation}
\begin{equation}
\scalebox{0.9}{$
C=\{c_1, c_2,...,c_N\}
$}
\end{equation}
Figure~\ref{fig:diversity in step-1} and \ref{fig:diversity in step-3} show that TeamLLM generally increases response diversity. This improvement stems from team role division and multi-phase collaboration, which introduce multiple perspectives and reduce homogenization from fixed thinking patterns of a single LLM \cite{liang-etal-2024-encouraging, lu2024llmdiscussionenhancingcreativity}. However, higher diversity does not always correspond to higher \textit{Flexibility} scores, consistent with existing findings in multi-LLM creativity studies \cite{lu2024llmdiscussionenhancingcreativity}.


\section{Conclusions}
We propose TeamLLM, a human-like team-oriented multi-LLM collaboration framework that enhances LLMs’ performance on contextualized multi-step tasks and construct the CGPST benchmark. Experiments results show that TeamLLM improves overall performance, particularly in divergent thinking and comprehensive problem solving, while fostering solution diversity and reducing output homogenization. Step-level and dimension-level analyses highlight that collaborative team roles enhance \textit{elaboration}, \textit{originality}, and \textit{focus}.


\section*{Limitations}
This work has several limitations. 

First, CGPST relies on fine-grained human evaluation across multiple steps and dimensions, which is time-consuming and limits scalability despite careful calibration. Future work will explore automated or hybrid evaluation methods to reduce annotation cost while preserving evaluation fidelity. 

Second, TeamLLM adopts a fixed set of team roles and a predefined collaboration protocol inspired by Belbin human team theory. While effective for contextualized multi-step tasks, the generalizability of this design to other task types and domains remains an open question.




\bibliography{custom}

\clearpage
\appendix

\section{Rationale for Team Roles Selection}
\label{app:role-selection}
In the late 1960s, Dr. Meredith Belbin concluded that team performance depends on the balance of team role distribution, which is known as the team role theory. His research showed that the most successful teams were made up of a diverse mix of behaviors \cite{Belbin2010, Belbin2022}. Inspired by this sociological and psychological theory, we attempt to introduce human team theory into multi-LLM teams for the first time, and based on this, propose \textbf{TeamLLM} framework. Belbin defined nine team roles, which can be further categorized into three types: \textit{thinking}, \textit{action}, and \textit{social}, as shown in Table~\ref{tab:belbin_roles}\footnote{\url{https://www.belbin.com/about/belbin-team-roles}}. In this appendix, we provide the rationale for selecting the four roles used in our framework: \textbf{Co-Ordinator}, \textbf{Plant}, \textbf{Monitor Evaluator}, and \textbf{Implementer}.

First, we exclude certain roles. The \textbf{Specialist} focuses on a specific domain and lacks cross-domain versatility, making it unsuitable for addressing the diverse problems that may arise in a team context. The \textbf{Resource Investigator} is responsible for exploring external opportunities. However, our study focuses on contextualized multi-step tasks like CGPST, which primarily evaluate the problem-solving abilities of LLMs, and does not require acquiring external resources.

Second, to ensure that the constructed team is representative and can maximally reflect the composition and collaboration patterns of human-like teams, we select roles covering Belbin’s three team role categories. From the \textit{thinking} category, we choose \textbf{Plant}, valuing its ability to “generate ideas and solve difficult problems.” From \textit{action}, we select \textbf{Implementer}, emphasizing its ability to “turn ideas into actions and organize work that needs to be done,” ensuring the team’s proposals are practical and feasible. From \textit{social}, we select \textbf{Co-Ordinator}, which “focuses on the team's objectives and delegates work appropriately,” making it more suitable than the \textbf{Teamworker} for completing the Task Initiation and Consensus Building phases.

Finally, since logical thinking is a critical aspect of problem-solving ability evaluation, we additionally include \textbf{Monitor Evaluator}, valuing its skill in “making impartial judgments where required and weighing up the team's options in a dispassionate way.” This prevents the team’s output from being merely practical but logically flawed. For example, generating an action plan that is feasible but contains internal contradictions in step dependencies. Furthermore, because CGPST emphasizes thinking, including two roles from the \textit{thinking} category is reasonable.

\begin{figure}[t]  
  \centering
  \includegraphics[width=\columnwidth]{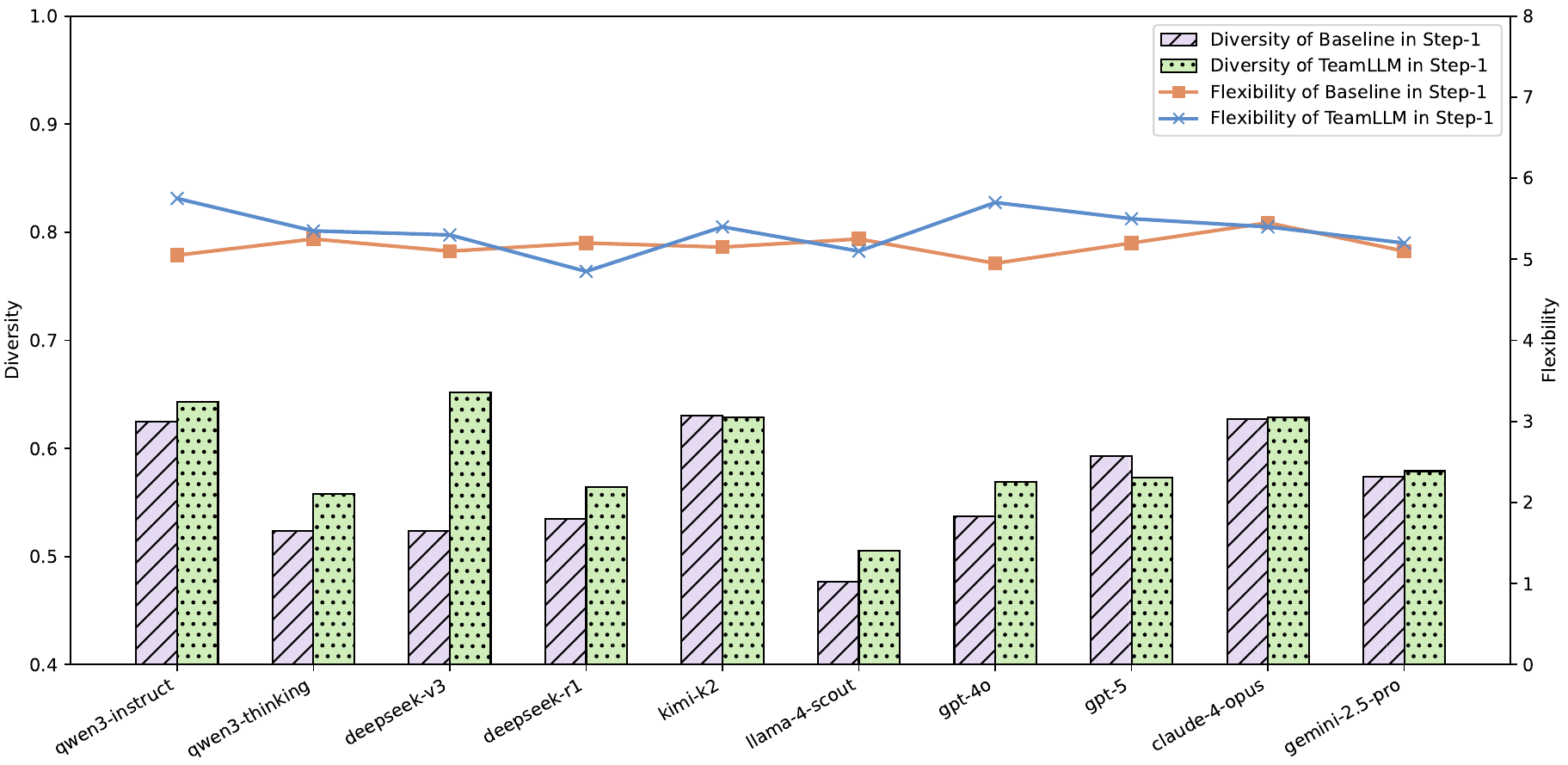}  
  \caption{Comparison of Diversity and Flexibility in Step-1.}
  \label{fig:diversity in step-1}
\end{figure}

\begin{figure}[t]  
  \centering
  \includegraphics[width=\columnwidth]{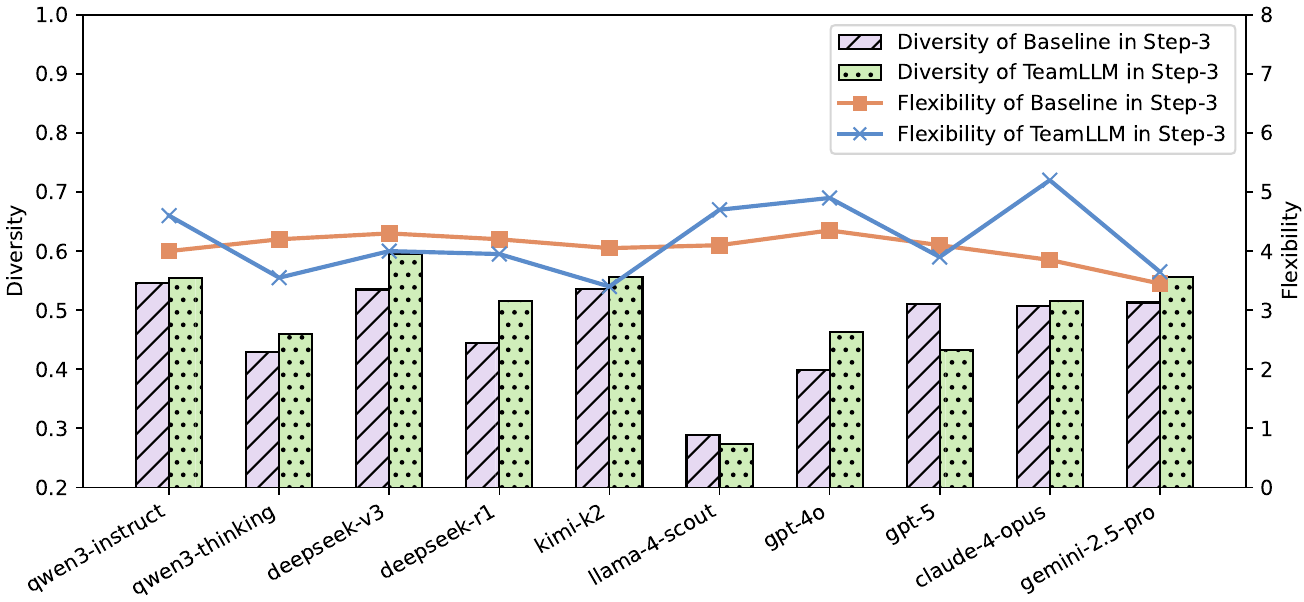}  
  \caption{Comparison of Diversity and Flexibility in Step-3.}
  \label{fig:diversity in step-3}
\end{figure}

\section{TeamLLM Prompts}
\label{app:TeamLLM Prompts Variables}
The prompt variables used in TeamLLM are listed in Table~\ref{tab:TeamLLM Prompts Variables}.

\section{Team Role Assignment}
\label{app:role-play}
In Team Formation, each LLM is assigned a distinct team role through a triplet \{\textit{Team\_Role}, \textit{Role\_Speciality}, \textit{Role\_Prompt}\}, which specifies the LLM’s and behavioral guidelines. This role-playing mechanism enables LLMs to propose, evaluate, execute, and coordinate solutions according to their role characteristics throughout the collaboration process, thereby simulating the division and interaction within human teams. Based on Belbin’s theory of team roles, the triplet prompts for each role are shown in Table~\ref{tab:team_roles} and the $Role\_Play\ prompt$ is shown in Table~\ref{tab:TeamLLM Prompts Variables}, serving as guidance for LLMs.

\section{CGPST Benchmark Step Prompt}
\label{app:CGPST step prompt}
Table~\ref{tab:CGPST step prompt} presents the detailed step-wise task information and corresponding output requirements for the CGPST benchmark. Each step specifies the task description and expected outputs, including guidelines for completion.

\section{Evaluation Rubrics for CGPST}
\label{app:rubrics}
To ensure a transparent and systematic evaluation of model performance on contextualized multi-step tasks, Table~\ref{tab:rubrics} provides the detailed rubrics for each step of the CGPST benchmark. The scoring criteria are designed to capture not only the correctness of responses but also the depth of thinking, creativity, and contextual alignment. Each step is assessed across multiple dimensions, with explicit descriptions of rubrics and corresponding maximum scores.

\section{Hyperparameter setting}
\label{app:hyperparameter}
on CGPST, different steps are designed to evaluate distinct abilities of LLMs. The \textit{temperature} affects performance: lower values yield more consistent outputs, while higher values promote diversity \cite{lu2024llmdiscussionenhancingcreativity}. To ensure fairness, we assign targeted temperature values for different team roles and experimental conditions, as shown in Table~\ref{tab:hyperparams}.

\begin{table}[htbp]  
\centering
{\fontsize{8.5pt}{8pt}\selectfont
\begin{tabular}{llc}
\toprule
\textbf{Hyper-parameter} & \textbf{Condition / Role} & \textbf{Value} \\
\midrule
\multirow{5}{*}{\textit{temperature}} 
 & TeamLLM – Co-Ordinator      & 0.5 \\
 & TeamLLM – Plant             & 0.8 \\
 & TeamLLM – Monitor Evaluator & 0.4 \\
 & TeamLLM – Implementer       & 0.4 \\
 & Baseline                   & 0.6 \\
\midrule
\textit{top-p} & – & 0.9 \\
\midrule
\textit{top-k} & – & 20 \\
\bottomrule
\end{tabular}
}
\caption{Hyper-parameter Settings}
\label{tab:hyperparams}
\end{table}

\section{Meta Prompt}
\label{app:meta prompt}
We use meta prompt to provide necessary background information about CGPST in the Team Formation phase. These prompts provide LLMs with essential context about the task’s future scenario, team roles, collaboration requirements, and necessary evaluation criteria. The meta prompts for TeamLLM framework and the baseline are presented in Table~\ref{tab:meta_prompts}, with {\textit{\{future scenario\}}} representing the future scenario for each task.

\section{Human Evaluation}
\label{app:human evaluation}
This study requires three experimental conditions: TeamLLM, the baseline and ablation condition, and collected a total of 300 responses based on 10 scenarios and 10 models. Given the inherently subjective nature of CGPST, we invited 15 trained human experts (H01-H15) to conduct double-blind scoring to ensure objectivity and reliability. Each response was independently scored by two unaware experts. First, all human experts were trained using a future scenario and a real model response, including detailed explanations of the evaluation dimensions for each step and instructions on how to use the \textbf{human scoring sheet} (see Appendix \ref{app:Human Evaluation Sheet}).

Before formal scoring, we carefully designed the scoring procedure. Each scoring session assigned a scenario to each expert, which included 10 responses from all models under the same condition within the same scenario, with the model names hidden. This arrangement ensured that experts only evaluated responses under the same scenario at a time, maintaining consistency and coherence. The scoring process consisted of 12 sessions in total, with each response taking approximately 25–30 minutes to evaluate. Each expert participated in about 3–5 sessions on average, reviewing roughly 30 responses.

As discussion in Section~\ref{section 5.3}, we employed \textbf{PCC} and \textbf{ICC} to ensure consistency. Considering that different steps and dimensions had varying score ranges, we normalized the scores before calculating correlations by dividing each score by the maximum score for the corresponding dimension. Specifically, for each pair of experts (Hx and Hy), PCC was calculated across the 10 responses in the same scenario under the same condition to assess agreement on individual ratings. At the same time, the 10 responses were treated as a whole to compute ICC, evaluating the stability and overall agreement of the two experts’ scores within the scenario.

A calibration mechanism was also introduced after scoring. When PCC for a response fell below a threshold of 0.7 (with a 5\% tolerance, set practically at 0.65), a third expert was invited to reassess that response. This mechanism ensured that the final scores were more stable and reliable. As shown in Table~\ref{tab:human_evaluation_consistency}, the calibration significantly improved scoring consistency.



\section{Dimension-level Results for each step}
\label{app:dimension-level rader}
\begin{figure*}[p]
    \centering
    \includegraphics[width=\textwidth,height=0.2\textheight,keepaspectratio]{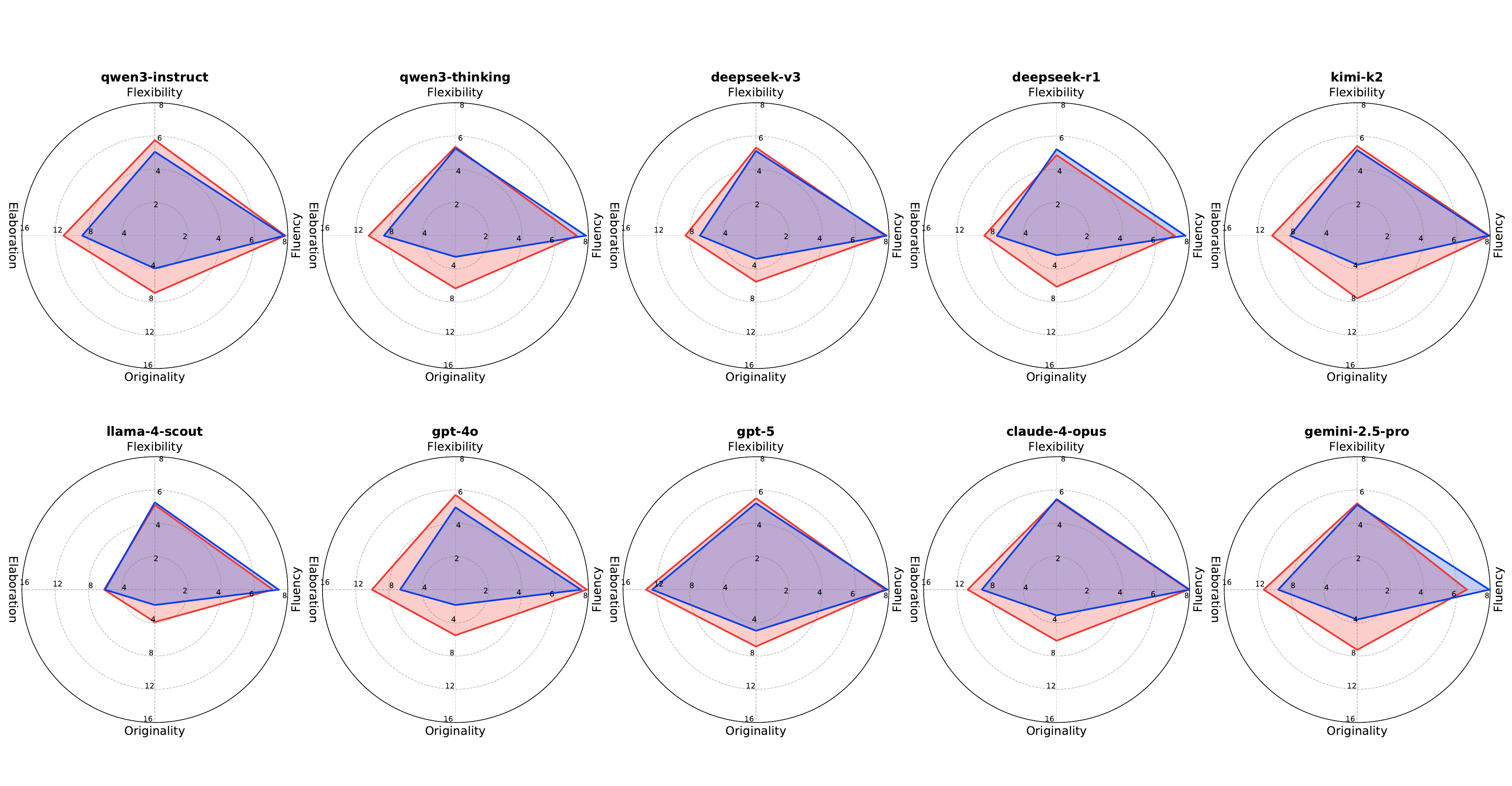} 
    \\[-1.2em] (a) Step-1

    \includegraphics[width=\textwidth,height=0.2\textheight,keepaspectratio]{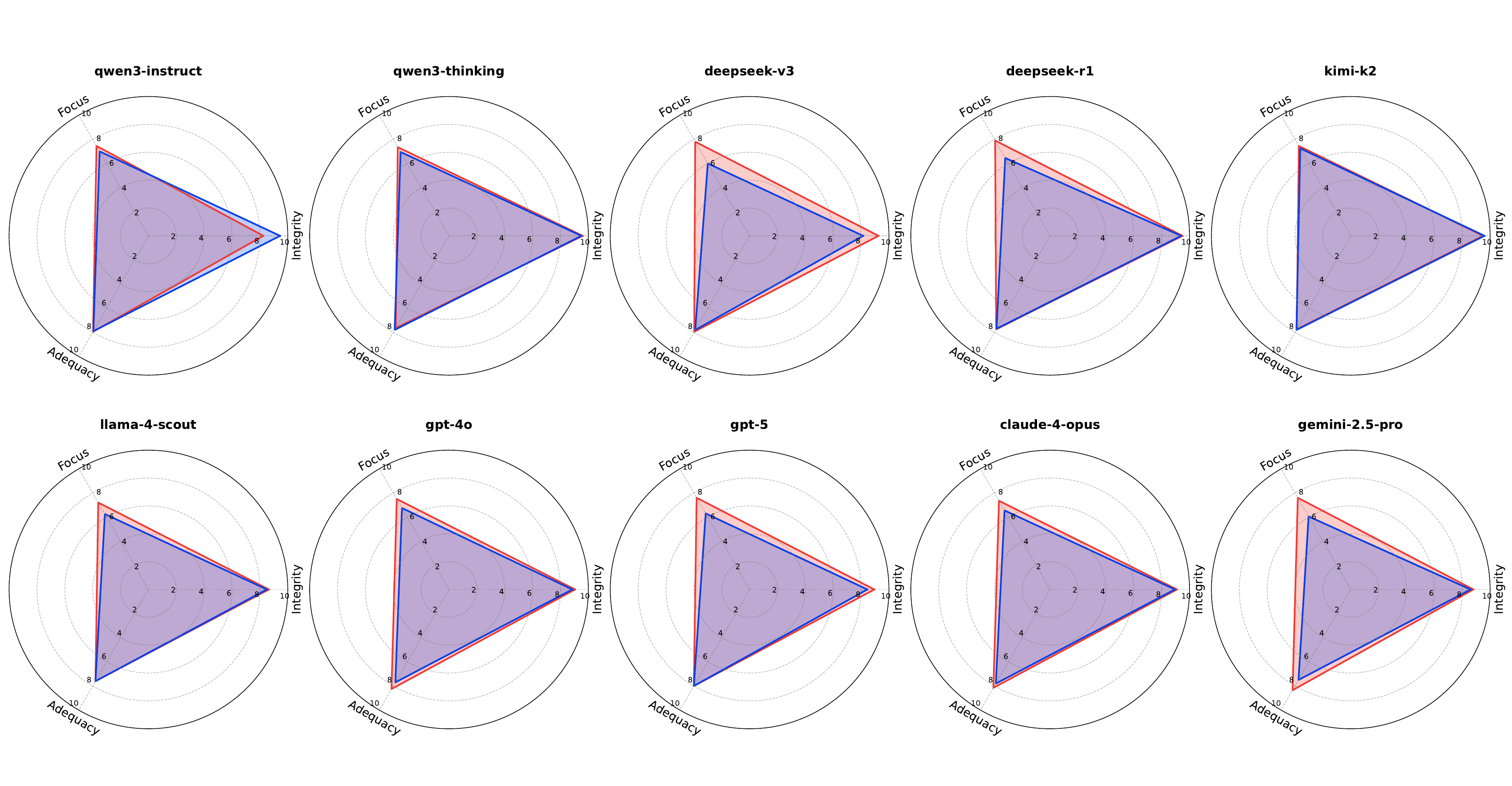} 
    \\[-1.2em] (b) Step-2

    \includegraphics[width=\textwidth,height=0.2\textheight,keepaspectratio]{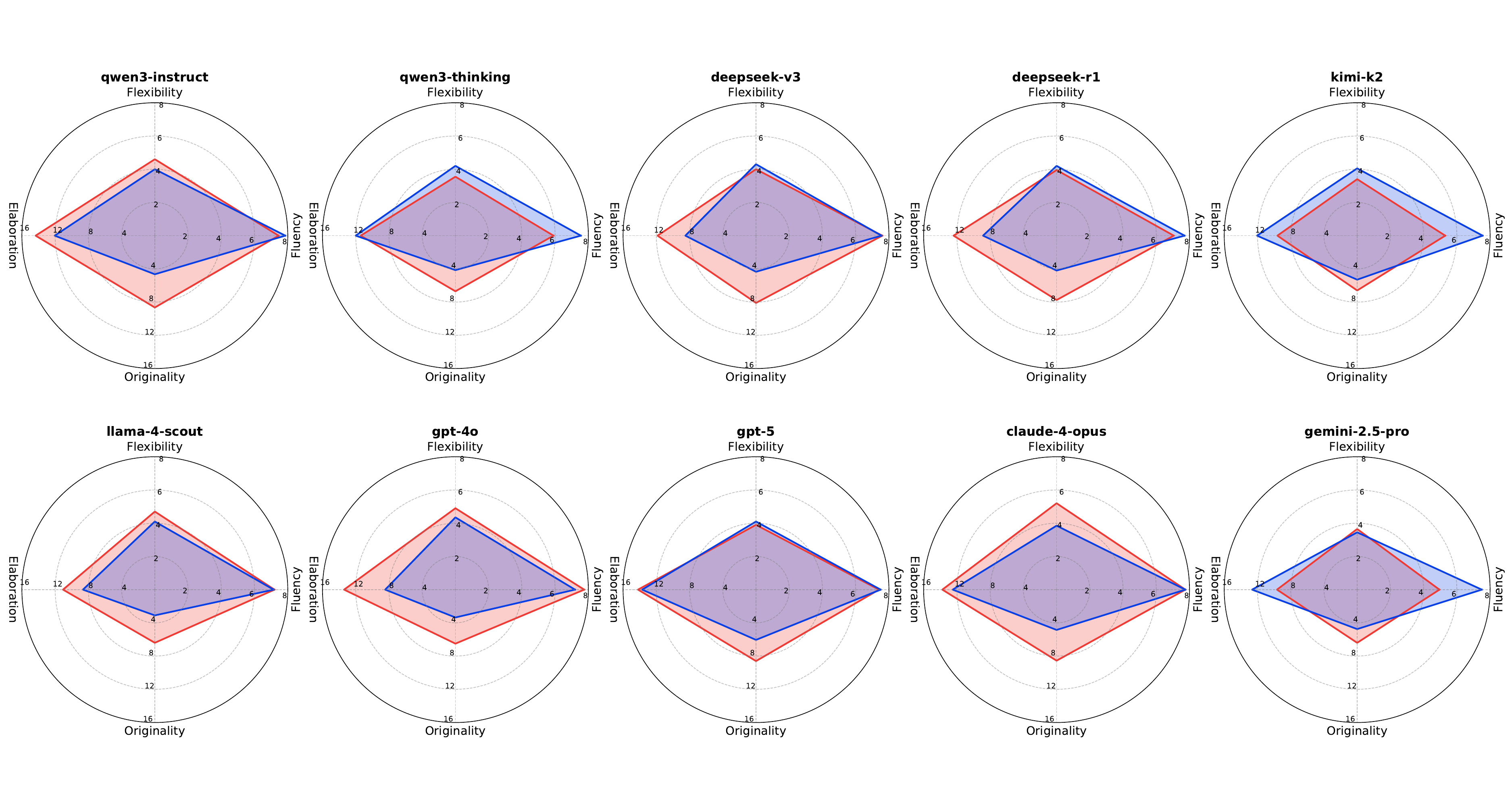} 
    \\[-1.2em] (c) Step-3

    \includegraphics[width=\textwidth,height=0.2\textheight,keepaspectratio]{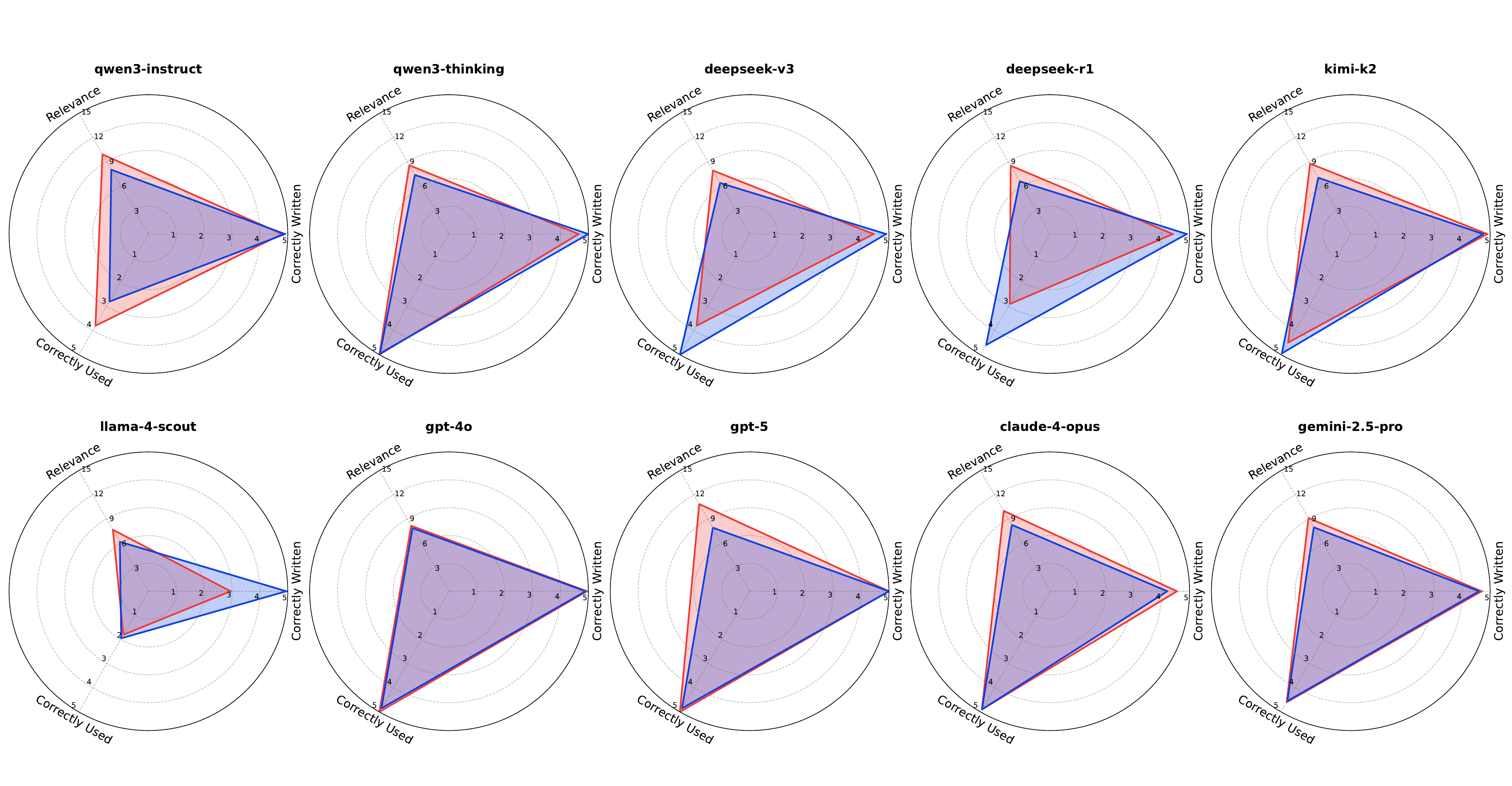} 
    \\[-1.2em] (d) Step-4 and Step-5

    \includegraphics[width=\textwidth,height=0.2\textheight,keepaspectratio]{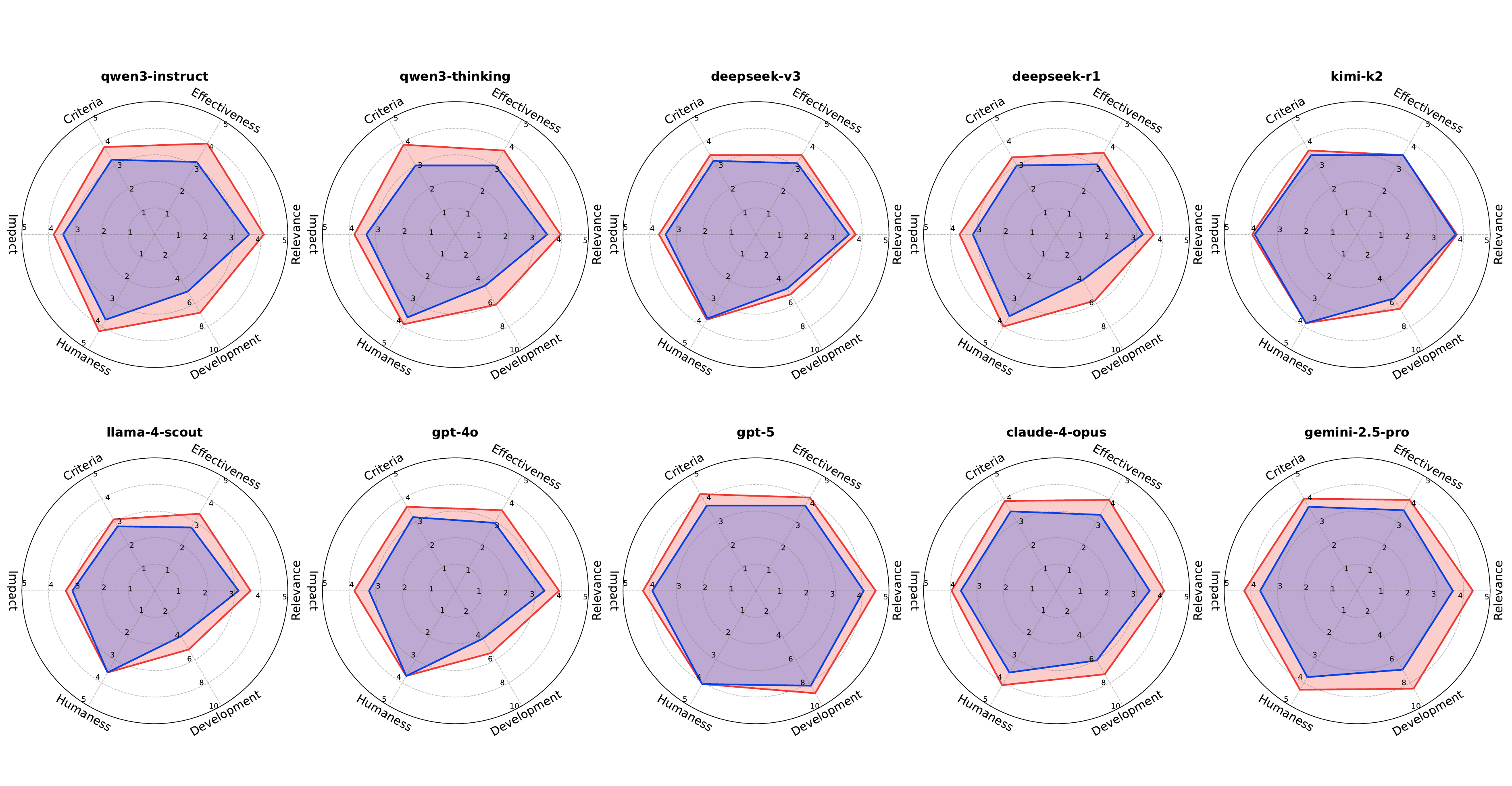} 
    \\[-1.2em] (e) Step-6

    \caption{Dimension-level performance of TeamLLM (red) and baseline (blue) across all steps.}
    \label{fig:dimension-level}
\end{figure*}

This appendix presents detailed radar charts of all models’ performance across all dimensions for each step on CGPST, as shown in Figure~\ref{fig:dimension-level}. These charts complement the step and dimension level analyses in the main text by providing a visual comparison of LLMs’ strengths and weaknesses in specific dimensions. Each figure illustrates the relative performance under both TeamLLM and the baseline, enabling a fine-grained examination of how team role-based collaboration affects individual dimension scores across all steps.

\begin{table}[t]
\centering
\footnotesize
\begin{tabular}{l c}
\toprule
\textbf{Type} & \textbf{Description} \\
\midrule
Perhaps & Ambiguous \\
Why             & Irrelevant to Scenario \\
Solution      & A Solution to a Challenge \\
Duplicate      & Similar to Another YES Challenge \\
Blank      & No Response Provided \\
\bottomrule
\end{tabular}
\caption{Types of invalid challenges in Step-1.}
\label{tab:Invalidity in Step-1}
\end{table}

\begin{table}[t]
\centering
\footnotesize
\begin{tabularx}{\linewidth}{l X} 
\toprule
\textbf{Type} & \textbf{Description} \\
\midrule
Perhaps & The relationship between the solution, the KVP, and the purpose is unclear. \\
Why & The solution is irrelevant to the underlying problem. \\
Duplicate & The solution is too similar to another Yes solution. \\
Blank & No response provided. \\
\bottomrule
\end{tabularx}
\caption{Types of invalid solutions in Step-3.}
\label{tab:Invalidity in Step-3}
\end{table}

\section{Human Evaluation Sheet}
\label{app:Human Evaluation Sheet}
\begin{figure*}[p]
    \centering
    \includegraphics[width=\textwidth,height=\textheight,keepaspectratio]{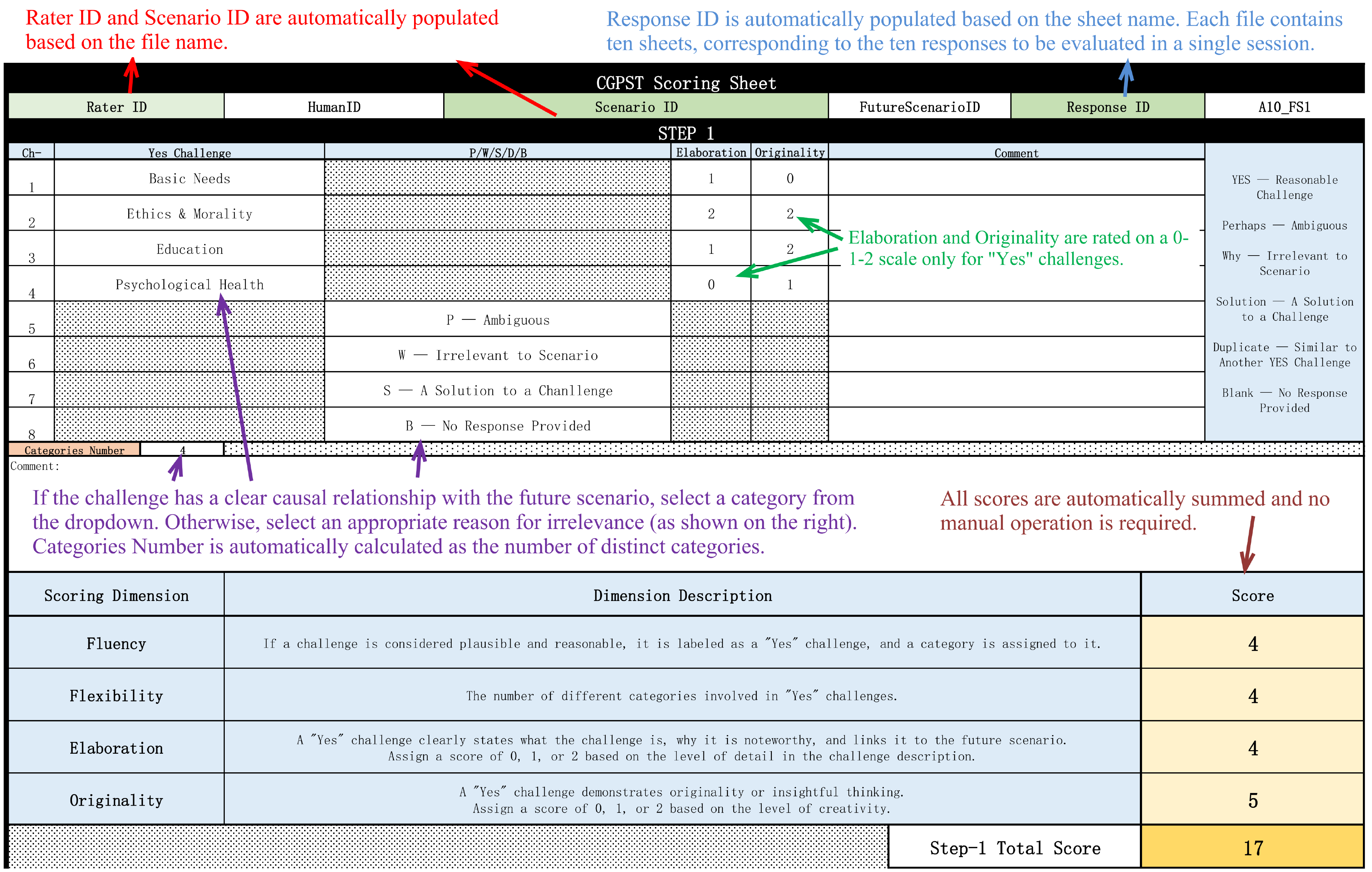} 
    \\[0.5em] (a) Step-1 Scoring Page

    \includegraphics[width=\textwidth,height=\textheight,keepaspectratio]{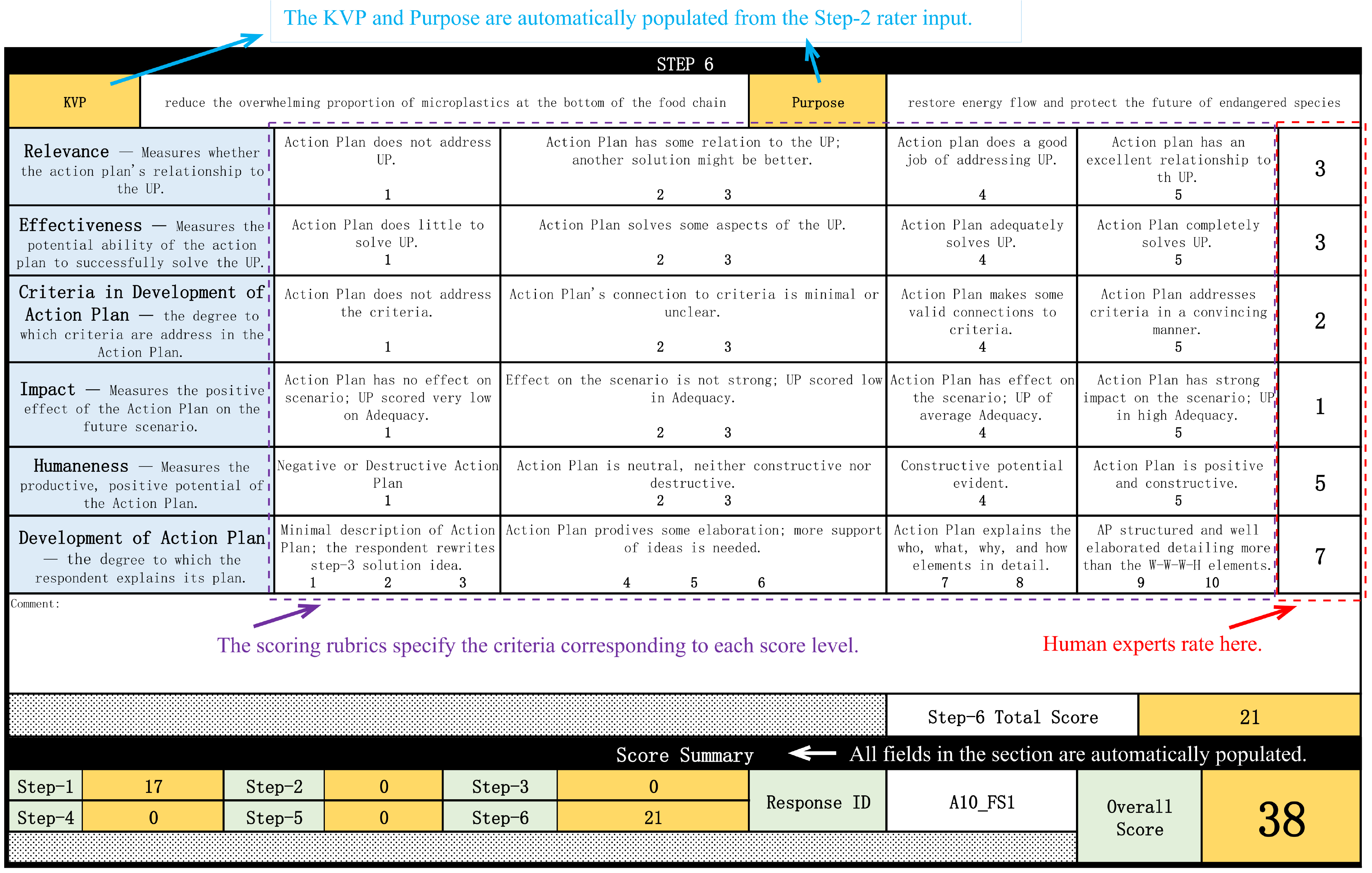} 
    \\[0.5em] (b) Step-6 Scoring Page

    \caption{Two representative pages of the Excel scoring sheet designed for human evaluation, with some annotated usage instructions.}
    \label{fig:human evaluation sheet}
\end{figure*}

For human evaluation, we designed a \textbf{dedicated Excel scoring sheet} for expert raters, adapted from the official FPSP scoring sheet\footnote{\url{https://resources.futureproblemsolving.org/article/how-evaluated-global-issues/}}. Additionally, we add cell decorations, cell locking, automatic calculations, and other built-in constraints in the score sheet to standardize the scoring process. The sheet also includes scoring rubrics that define what each score level (e.g., 0, 1, and 2) represents for each criterion, helping raters calibrate their judgments. Two representative pages of the evaluation sheet is presented in Figure \ref{fig:human evaluation sheet}, with some annotations indicating usage instructions.

\section{Step 5 Answer Instances of \textit{llama-4-scout} Model}
\label{app:step-5 instances of llama}
Table~\ref{tab:step-5 instances of llama} shows some representative examples of \textit{llama-4-scout} in Step-5. Among the 30 responses generated by \textit{llama-4-scout}, 24 contained various ranking errors, 2 failed to follow the stepwise requirements for answering, and only 4 were correct. This indicates that \textit{llama-4-scout} exhibits poor performance in terms of logical ranking.

\section{Complete CGPST Example}
\label{app:complete CGPST example}
For illustrating the complete CGPST process and human scoring, we provide a complete CGPST scenario (FS10) as shown in Table~\ref{tab:CGPST_scenario_example}, along with the full six-step responses generated by the \texttt{kimi-k2-0711-preview} under the TeamLLM condition, as shown in Table \ref{tab:Step-1 of the CGPST Example} to \ref{tab:Score Summary of the CGPST Example}. For each step, we also include scores provided by two human experts. The dataset identifier for this response is A05\_FS10, and the two human experts are identified as H01 and H02. The PCC agreement between the two experts for this response is 0.8534, which exceeds the predefined consistency threshold. Therefore, no post-hoc calibration is applied to the assigned scores. Where necessary, we include additional explanations to help readers better understand the scoring process.

\section{Dimension Details for Step-1 and Step-3 of the CGPST}
\label{app:dimension details for step-1 and Step-3}
As described in Section~\ref{Section 4.2}, Step-1 and Step-3 of the CGPST focus on divergent thinking. Consistent with prior research on divergent thinking \cite{10.1145/3613904.3642731, 10.1145/3706598.3714198, Zhao2025, lu2024llmdiscussionenhancingcreativity}, we adopt the four dimensions from the Torrance Tests of Creative Thinking (TTCT) \cite{torrance1966torrance} assessment to evaluate these steps. Here, we provide the necessary details for each of the four dimensions.
\begin{itemize}
  \item \textbf{Fluency}: the sheer volume of meaningful ideas produced in reaction to a given stimulus. In proposed CGPST benchmark, \textit{fluency} is measured by counting the number of valid challenges or solutions proposed by the model. Specifically, as shown in Table \ref{tab:rubrics}, in Step 1, a challenge is considered valid only if it clearly reflects a causal relationship with the future scenario. In Step 3, a solution is considered valid only if it addresses the underlying problem identified in Step 2. The final \textit{fluency} score is calculated based on the total number of valid challenges or solutions. For invalid challenges or solutions, we defined several types of invalidity for human raters to choose from, as shown in Table \ref{tab:Invalidity in Step-1} and Table \ref{tab:Invalidity in Step-3}.
  \item \textbf{Flexibility}: the diversity of categories within the responses. Given that the Future Problem Solving Program (FPSP) is a mature problem-solving program with over 50 years \cite{treffinger2012four}, we adopt the 20 categories defined in the program. Table \ref{tab:Category List} lists all categories and their descriptions.
  \item \textbf{Elaboration}: the depth or granularity of details within the responses. Specifically, this refers to the level of detail in describing the causal relationship between the challenge and the future scenario in Step-1, or the extent to which a solution elaborates on how it addresses the identified underlying problem in Step-3.
  \item \textbf{Originality}: the uniqueness or novelty of answers. As described in Appendix \ref{app:human evaluation}, during each human evaluation session, each expert is assigned all model responses under the same scenario and condition, resulting in a total of ten responses per session. That is, each expert evaluates responses generated for a single future scenario within one session. Prior to scoring, experts carefully read the scenario and develop an initial understanding of the context. Based on this understanding, and through horizontal comparisons across the challenges or solutions proposed by different models, experts assign scores for the \textit{originality} dimension. This evaluation design helps reduce subjectivity and improves the reliability and accuracy of the \textit{originality} assessment.
\end{itemize}

\begin{table*}[htbp]
\centering
\begin{tabular}{p{5cm} p{9cm}}
\toprule
\textbf{Prompts Variables} & \textbf{Content} \\
\midrule
$Role\_Play \ prompt$

&

Your role in the team is \{Team\_Role\}, with the specialty of \{Role\_Speciality\}. The other team members are \{Other\_Members\}, each performing different team roles and responsibilities. \{Role\_Prompt\} \\
\cdashline{1-2}
$warm\_up \ prompt$

&

Before starting the task, please guide the team members through a warm-up interaction to help them get to know each other and foster a collaborative atmosphere. Encourage each member to briefly introduce their role and responsibilities and greet everyone. Use a natural and friendly tone to open the session, inviting members to speak one by one to complete this warm-up activity. Please note that this stage should focus only on team warm-up and not involve the task itself. \\
\cdashline{1-2}
$Task\_Initiation\ prompt$

&

The information for Step \{Step\_Number\} is as follows: 

\vspace{1em}
\{Step\_Name\}, \{Step\_Description\}, \{Step\_Output\}

\vspace{1em}
Please clearly and concisely communicate the information to the team, encouraging members to leverage their strengths to complete it together. Note that you do not need to assign specific roles; just convey the task information. \\
\cdashline{1-2}
$Consensus\_Building\ prompt$

&

The discussion for Step \{Step\_Number\} has concluded. Based on the contributions of the team, synthesize their opinions and viewpoints to produce the final answer for this step. Note that the final answer should meet the requirements in [Step\_Output]. Please provide the answer directly without any explanation. \\
\bottomrule
\end{tabular}
\caption{Prompts variables in TeamLLM.}
\label{tab:TeamLLM Prompts Variables}
\end{table*}

\begin{table*}[htbp]
\centering
\footnotesize
\begin{tabular}{p{1.5cm} p{2.7cm} p{5.2cm} p{4.5cm}}
\toprule
Category & Roles & Description & Strengths \\
\midrule
Thinking & Plant & Tends to be highly creative and good at solving problems in unconventional ways. & Creative, imaginative, free-thinking, generates ideas and solves difficult problems. \\
~ & Monitor Evaluator & Provides a logical eye, making impartial judgements where required and weighs up the team's options in a dispassionate way. & Sober, strategic and discerning. Sees all options and judges accurately. \\
~ & Specialist & Brings in-depth knowledge of a key area to the team. & Single-minded, self-starting and dedicated. They provide specialist knowledge and skills. \\
\cdashline{1-4}
~Action & Shaper & Provides the necessary drive to ensure that the team keeps moving and does not lose focus or momentum. & Challenging, dynamic, thrives on pressure. Has the drive and courage ~to overcome obstacles. \\
~ & Implementer & Needed to plan a workable strategy and carry it out as efficiently as possible. & Practical, reliable, efficient. Turns ideas into actions and organises work that needs to be done. \\
~ & Completer Finisher & Most effectively used at the end of tasks to polish and scrutinise the work for errors, subjecting it to the highest standards of quality control. & Painstaking, conscientious, anxious. Searches out errors. Polishes and perfects. \\
\cdashline{1-4}
Social & 
Resource Investigator & Uses their inquisitive nature to find ideas to bring back to the team.  & Outgoing, enthusiastic. Explores opportunities and develops contacts. \\
~ & Co-ordinator & Needed to focus on the team's objectives, draw out team members and delegate work appropriately. & Mature, confident, identifies talent. Clarifies goals. \\
~ & Teamworker & Helps the team to gel, using their versatility to identify the work required and complete it on behalf of the team. & Co-operative, Co-operative, perceptive and diplomatic. Listens and averts friction. \\
\bottomrule
\end{tabular}
\caption{Belbin Team Roles.} %
\label{tab:belbin_roles}
\end{table*}

\begin{table*}[htbp]
\centering
\begin{tabular}{p{7.5cm} p{7.5cm}}
\toprule
\textbf{TeamLLM} & \textbf{Baseline} \\
\midrule
\hspace{2em}You are about to participate in a contextualized task. In this task, you will receive a scenario about future societal issues. Please work with your team to complete the problem solving task according to your assigned role and responsibilities.

\hspace{2em}You will participate as a team member in all six steps of the problem solving process. Teamwork is key to success! You are expected to actively listen to others’ ideas, clearly express your own viewpoints, and collaborate based on your role, engaging in a systematic discussion throughout the task. Please maintain an active dialogue, continuously communicate and coordinate with your teammates, and work together to complete the task.

\vspace{1em}
\hspace{2em}The future scenario for this task is as follows:

\vspace{1em}
\hspace{2em}\textit{\{future scenario\}}

\vspace{1em}
\hspace{2em}Please carefully read the above scenario and use it as the basis for participating in the subsequent problem solving task.

\vspace{1em}
\hspace{2em}Please note that during the task, you must adhere to the following requirements:
\vspace{1em}

\hspace{2em}1. Actively collaborate with your team members, carefully consider their contributions, and work together to advance the task.

\vspace{1em}
\hspace{2em}2. Clearly understand and adhere to your assigned team role and responsibilities, ensuring consistency in role-playing.

\vspace{1em}
\hspace{2em}3. Follow the six-step problem solving process strictly.

\vspace{1em}
\hspace{2em}4. All responses should be in Chinese, with clear language and accurate logic.

\vspace{1em}
\hspace{2em}5. For each response, only output what you intend to say—do not include any role descriptions, labels, or guiding text.
&
\hspace{2em}You are about to participate in a contextualized task. In this task, you will receive a scenario about future societal issues. Please follow the task steps and independently complete the task.
\vspace{1em}

\hspace{2em}The future scenario for this task is as follows:

\vspace{1em}
\hspace{2em}\textit{\{future scenario\}}

\vspace{1em}
\hspace{2em}Please carefully read the above scenario and use it as the basis for participating in the subsequent problem solving task.

\vspace{1em}
\hspace{2em}Please note that during the task, you must adhere to the following requirements:

\vspace{1em}
\hspace{2em}1. For each response, only output what you intend to say—do not include any explanations, labels, or guiding text.

\vspace{1em}
\hspace{2em}2. All responses should be in Chinese, with clear language and accurate logic. \\
\bottomrule
\end{tabular}
\caption{Meta Prompts for TeamLLM and Baseline Conditions.}
\label{tab:meta_prompts}
\end{table*}

\begin{table*}[t]
\centering
\begin{tabular}{@{}l p{0.2\textwidth} p{0.6\textwidth}@{}}
\toprule
\textbf{Team\_Role} & \textbf{Role\_Speciality} & \textbf{Role\_Prompt} \\ 
\midrule
Co-Ordinator & Team guidance, task organization, consensus integration & As the Co-Ordinator of the team, your primary responsibility is to organize and guide team collaboration, drive task progress, and integrate perspectives. Your contributions should demonstrate leadership and structure, clearly guiding the team to focus on the current step, engaging each member to fulfill their role, and synthesizing everyone’s response to form team consensus. \\ 

Plant & Innovative thinking, generating novel solutions, proposing unusual yet potentially valuable ideas & As the Plant of the team, your main responsibility is to propose creative, even unconventional ideas to inspire the team. Your contributions should be imaginative and open-minded, daring to challenge conventions. Focus on the current task and generate as many creative ideas as possible to provide a draft for team discussion. \\ 

Monitor Evaluator & Rational analysis, logical judgment, reasoning & As the Monitor Evaluator of the team, your main responsibility is to evaluate the Plant’s contributions from a rational and logical perspective. Focus on the current task step, applying your reasoning and analytical skills to assess the feasibility and logical consistency of the Plant’s ideas, providing guidance for the team’s final integrated answer. \\ 

Implementer & Task execution, feasibility analysis, pragmatic implementation & As the Implementer of the team, your main responsibility is to assess the Plant’s ideas from a feasibility and practical implementation perspective. Focus on the current task step, leveraging your expertise in pragmatism and execution to provide recommendations that ensure the team’s final integrated answer is actionable and realistic. \\ 
\bottomrule
\end{tabular}
\caption{Role play prompts. Each role is defined as a triplet \{Team\_Role, Role\_Speciality, Role\_Prompt\} guiding the LLM’s behavior.}
\label{tab:team_roles}
\end{table*}

\begin{table*}
\centering
\begin{tabular}{>{\small}p{2cm}>{\small}p{3cm}>{\small}p{10.15cm}}
\toprule
\textbf{Step\_Name} & \textbf{Step\_Description} & \textbf{Step\_Output} \\
\midrule
Step-1: Identify Challenges &
Please carefully read the future scenario and identify up to eight challenges or problems specifically related to the scenario context. &
Please describe each challenge in a complete sentence, following these requirements:
\begin{enumerate}[itemsep=0pt, parsep=0pt, topsep=1pt]
    \item Use declarative sentences.
    \item Use modal verbs like “might,” “could,” or “should.”
    \item Explain what the challenge is, why it is a challenge, and how it connects to the scenario.
    \item Number each challenge, e.g., “1. xxx.”
\end{enumerate} \\

Step-2: Select an Underlying Problem &
From the challenges in Step-1, select the most impactful one and reformulate it into a focused core problem statement. &
Provide a complete description including:
\begin{enumerate}[itemsep=0pt, parsep=0pt, topsep=1pt]
    \item Challenge number: the identifier of the specific challenge from Step 1 that is being developed into the underlying problem.
    \item Conditional phrase (CP): a fact or condition drawn from the future scenario, which provides the theoretical or situational basis for the problem.
    \item Stem + Key Verb Phrase (KVP): the core phrasing of the underlying problem, usually starting with “How might we…” or “In what ways can we…”. The KVP should contain only one active verb specifying the main action or intervention to be taken, and should avoid absolute or overly broad verbs to ensure focus and feasibility.
    \item Purpose: typically expressed with “in order to” or “so that”, clarifying the intended goal of the KVP.
    \item Future scenario parameters: the three parameters of time, location, and theme that situate the underlying problem within the scenario.
\end{enumerate} \\

Step-3: Produce Solutions &
Generate up to eight possible solutions based on the underlying problem. &
Each solution should:
\begin{enumerate}[itemsep=0pt, parsep=0pt, topsep=1pt]
    \item Each solution must be written as a complete sentence.
    \item Use “will” rather than “might” to indicate certainty.
    \item Each solution should address at least three of the following aspects: Who, What, How, Why, When, and Where.
    \item Ensure alignment with the key verb phrase (KVP) and the intended purpose of the underlying problem.
    \item Begin each solution with a number, e.g., “1. …”.
\end{enumerate} \\

Step-4: Select Criteria &
Create five criteria to evaluate the solutions. &
Each criterion should:
\begin{enumerate}[itemsep=0pt, parsep=0pt, topsep=1pt]
    \item Be properly phrased: single dimension, superlatives as needed, indicate evaluation direction, phrased as a question
    \item Be relevant to the underlying problem
    \item Numbered, e.g., “1. xxx”
\end{enumerate} \\

Step-5: Apply Criteria to Top Solution & Evaluate the eight solutions from Step-3 using the criteria from Step-4 in a matrix format. &
Please provide the answers for this step in the following matrix (grid) format:

\texttt{Solution ID | Criterion 1 | Criterion 2 | Criterion 3 | Criterion 4 | Criterion 5 | Total Score}

\texttt{1       |   5    |   7    |   6    |   4    |     8     |  30}

\texttt{2       |   8    |   6    |   7    |   5    |     7     |  33}

\texttt{...     |  ...   |  ...   |  ...   |  ...   |    ...    | ...}

Highest-scoring solution ID: <ID>, Solution: <content of the highest-scoring solution>

Requirements:
\begin{enumerate}[itemsep=0pt, parsep=0pt, topsep=1pt]
    \item For each criterion, all solutions must be scored.
    \item Scores for each criterion should range from 1 to $x$, where 1 represents the worst-performing solution and $x$ represents the best.
    \item No two solutions may receive the same score under the same criterion; i.e., each column must be a unique permutation of 1 to $x$.
    \item Provide both the full scoring matrix and the ID and content of the highest-scoring solution.
\end{enumerate} \\

Step-6: Develop an Action Plan &
Develop the top solution from Step-5 into an actionable plan. &
Develop the highest-scoring solution selected in Step-5 into a comprehensive action plan. The plan should systematically and thoroughly explain how the underlying problem is addressed, how the solution will be implemented, and how the future scenario will be impacted. Please provide the final action plan in the following format:

\hspace*{2em}Best Solution ID: \# (fill in the solution ID)

\hspace*{2em}Best Solution Content: (fill in the content of the best solution)

\hspace*{2em}Action Plan: (full, detailed action plan) \\

\bottomrule
\end{tabular}
\caption{Step-wise Task Requirements and Outputs on CGPST.}
\label{tab:CGPST step prompt}
\end{table*}

\begin{table*}
\centering
\begin{tabular}{>{\normalsize}p{3cm}>{\normalsize}p{3cm}>{\normalsize}p{8cm}>{\normalsize}p{1cm}}
\toprule
\textbf{Step} & \textbf{Dimension} & \textbf{Rubric} & \textbf{Max Score} \\
\midrule
\multirow[t]{4}{3cm}{Step-1: Identify Challenges} 
& Fluency & Each reasonable challenge is counted as \textit{Yes}; one \textit{Yes} challenge = 1 point. & 8 \\
& Flexibility & If a \textit{Yes} challenge belongs to a new category, 1 point is awarded. & 8 \\
& Elaboration & If a \textit{Yes} challenge has a clear logical-causal relationship and is described in detail, 2 points are awarded. & 16 \\
& Originality & If a \textit{Yes} challenge demonstrates unique creativity, 2 points are awarded. & 16 \\

\multirow[t]{6}{3cm}{Step-2: Select an Underlying Problem} 
& Integrity: CP & Whether it is connected to the future scene and KVP. & 2 \\
& Integrity: Stem \& KVP & Contains only one clear active verb phrase. & 3 \\
& Integrity: Purpose & Contains only one purpose that is associated with KVP. & 3 \\
& Integrity: Future Scene Parameters & Clearly reflects the theme, location, and time. & 2 \\
& Focus of Underlying Problem & High-quality KVP, clear purpose, strong linkage between KVP and purpose, and relevant to the future scene. & 10 \\
& Adequacy / Importance & The problem is major and important in the future scenario. & 10 \\

\multirow[t]{4}{3cm}{Step-3: Produce Solutions} 
& Fluency & A solution idea that addresses the KVP and supports the action purpose is marked as \textit{Yes}. Each \textit{Yes} counts for 1 point. & 8 \\
& Flexibility & A \textit{Yes} solution that belongs to a new category counts for 1 point. & 8 \\
& Elaboration & A \textit{Yes} solution that includes three elements out of who, what, how, why, when, counts for 2 points. & 16 \\
& Originality & A \textit{Yes} solution that demonstrates unique creativity counts for 2 points. & 16 \\

\multirow[t]{2}{3cm}{Step-4: Select Criteria} 
& Correctly Written & A criterion is written correctly, 1 point each. & 5 \\
& Relevance & Criteria are clear, specific, and highly relevant to the underlying problem, 3 points each. & 15 \\

Step-5: Apply Criteria to Top Solution 
& Correctly Used & The scoring matrix is used correctly. & 5 \\

\multirow[t]{6}{3cm}{Step-6: Develop an Action Plan} 
& Relevance & The action plan is highly relevant to the underlying problem. & 5 \\
& Effectiveness & The action plan effectively and comprehensively addresses the underlying problem. & 5 \\
& Criteria in Development of Action Plane & The action plan adequately responds to the established criteria. & 5 \\
& Impact & The action plan has a significant and positive impact on the future scenario. & 5 \\
& Humaneness & The action plan demonstrates clear human concern and empathy. & 5 \\
& Development & The action plan is clearly structured, sufficiently elaborated, and highly feasible and complete. & 10 \\
\bottomrule
\end{tabular}
\caption{Evaluation Rubrics for CGPST Benchmark.}
\label{tab:rubrics}
\end{table*}

\begin{table*}[t]
\centering
\renewcommand{\arraystretch}{1.5}
\begin{tabular}{
>{\raggedright\arraybackslash}p{0.16\textwidth}
>{\raggedright\arraybackslash}p{0.394\textwidth}
>{\raggedright\arraybackslash}p{0.393\textwidth}
}

\rowcolor{teal!30}
\multicolumn{3}{>{\raggedright\arraybackslash}p{\textwidth}}{\textbf{Ranking errors (A06\_FS1):} The ranking scores assigned by each criteria are not a permutation of 1 to the total number of solutions (i.e., 8).} \\

\rowcolor{teal!10}
\multicolumn{3}{>{\raggedright\arraybackslash}p{\textwidth}}{{\texttt{Solution ID | Criterion 1 | Criterion 2 | Criterion 3 | Criterion 4 | Criterion 5 | Total Score}

\texttt{| 1    | 8    | 6    | 7    | 9    | 5    | 35 |}

\texttt{| 2    | 5    | 8    | 9    | 7    | 8    | 37 |}

\texttt{| 3    | 9    | 5    | 6    | 8    | 9    | 37 |}

\texttt{| 4    | 6    | 9    | 8    | 6    | 7    | 36 |}

\texttt{| 5    | 7    | 7    | 5    | 9    | 6    | 34 |}

\texttt{| 6    | 4    | 4    | 9    | 8    | 8    | 33 |}

\texttt{| 7    | 3    | 9    | 7    | 5    | 9    | 33 |}

\texttt{| 8    | 2    | 6    | 8    | 7    | 5    | 28 |}} 

\vspace{1em}
The solution with the highest total score is: 2. 

The solution is: The government will formulate and enforce strict cybersecurity regulations, requiring companies developing autonomous driving systems to comply with specific safety standards. Regular security audits and evaluations will be conducted on these systems, with full implementation scheduled to be completed in Istanbul by 2052.} \\

\multicolumn{3}{p{\textwidth}}{~} \\

\rowcolor{teal!30}
\multicolumn{3}{>{\raggedright\arraybackslash}p{\textwidth}}{\textbf{Ranking errors (A06\_FS5):} The ranking matrix is incomplete.} \\

\rowcolor{teal!10}
\multicolumn{3}{>{\raggedright\arraybackslash}p{\textwidth}}{{\texttt{Solution ID | Criterion 1 | Criterion 2 | Criterion 3 | Criterion 4 | Criterion 5 | Total Score}

\texttt{| 9    | 9    | 9    | 9    | 9    | 9    | 45 |}} 

\vspace{1em}
The solution with the highest total score is: 9. 

The solution is: By leveraging technologies such as artificial intelligence, blockchain, and the Internet of Things (IoT), a decentralized platform for cherry tree breeding and disease resistance monitoring will be established. This platform can connect cherry farmers, research institutions, and government departments in Tasmania, enabling data sharing, resource integration, and collaborative work. It will provide functions including real-time monitoring, predictive analysis, and smart contract-based incentives, so as to accelerate the cultivation of disease-resistant cherry tree varieties and improve the risk resilience of Tasmania’s cherry industry.} \\

\multicolumn{3}{p{\textwidth}}{~} \\

\rowcolor{teal!30}
\multicolumn{3}{>{\raggedright\arraybackslash}p{\textwidth}}{\textbf{Non-compliant responses (A06\_FS4):} The model fails to apply five criteria generated in Step-4 to rank the solutions with the scoring matrix.} \\

\rowcolor{teal!10}
\multicolumn{3}{>{\raggedright\arraybackslash}p{\textwidth}}{The solution with the highest total score is: 2. 

The solution is: Local schools will collaborate with communities to launch a toy and book sharing program. A toy and book exchange center will be established, where residents can donate and borrow items. The program is scheduled to be fully completed within three months and will be jointly maintained by teachers and students of the participating schools.} \\

\end{tabular}

\caption{Step 5 Answer Instances of \textit{llama-4-scout}.}
\label{tab:step-5 instances of llama}
\end{table*}

\begin{table*}[t]
\centering
\renewcommand{\arraystretch}{1.5}
\begin{tabular}{>{\raggedright\arraybackslash}p{\textwidth}}

\rowcolor{teal!30}
\textbf{FS10: Ocean Soup Future Scenario} \\

\rowcolor{teal!10}
\hspace{2em}As Jobie Sakai leans on the railing of the Ola Kai, she admires the beauty of her ocean paradise. Jobie, a fifth generation Hawaiian devoted to the future of her homeland, is dedicated to her job as an environmental chemist aboard the Ola Kai \#6, one of Hawai'i's floating science laboratories. The Ola Kai (meaning "healthy ocean") Project is a combined effort of the Hawaiian Environmental Council and the University of Hawai'i.

\vspace{1em}
\hspace{2em}Now, in 2035 after 15 active years, the project is struggling to live up to its nickname: the OK Project. Originally, the OK Project focused on the waters affected by Hawai'i's island-generated pollutants. Public interest in the project led to a resurgence in eco-education; recycling and the reduced use of plastics became an accepted part of island life for Hawai'i residents. The "adopt a beach" clean-up program became a popular draw for eco-tourists. However, researchers like Jobie became increasingly aware that their efforts were not enough.

\vspace{1em}
\hspace{2em}The world's largest manufacturers of plastic products border both sides of the Pacific. A ten million square mile system of rotating currents called the North Pacific Gyre has its axis near the 137 islands of the Hawaiian chain. Pacific environmental regulations have historically been weak or disregarded by heavy industrial nations who continue to use these waters as a dumping ground. Consequently, the 1500 mile-long archipelago paradise has been attacked by ocean soup for many years.

\vspace{1em}
\hspace{2em}The soup surrounds Hawai'i, placing the islands and their resources at risk of permanent damage. Especially vulnerable are the sparsely-inhabited northwest islands, the world's largest protected marine sanctuary that is home to many endangered fish, birds, seals, and Hawai'i's beleaguered fishing industry. Eco-tourism routes have been altered to reduce impact on indigenous species and circumnavigated due to the location of floating laboratories.

\vspace{1em}
\hspace{2em}With the increasing damaging effects from ocean soup on the island chain, Jobie and her coworkers realized that the Ola Kai Project's numerous floating labs had to broaden their territory while narrowing their focus. Due to the scope of the damage, the project directors reached out to other groups working in the Pacific and consulted with the National Oceanic and Atmospheric Administration (NOAA). It was determined that the best approach would be to divide up the responsibilities among agencies. Now the OK Project labs focus solely on the battle against microplastics, leaving the collection of larger trash to other organizations.

\vspace{1em}
\hspace{2em}Ola Kai laboratory crews record various data for analysis, keeping track of multiple fish species and beneficial organisms like plankton. Specified lab crews weigh the microplastic debris collected on a weekly basis and compile that data while Jobie and other chemists continue their in-depth examination of the plastic degradation and its effect on the waters surrounding the islands.

\vspace{1em}
\hspace{2em}Collection of debris that is smaller than a pencil eraser has often done more harm than good to sea life. After experimenting with several collection methods, the OK Project currently uses below-surface robotic collectors that randomly collect plastic particles being carried by the currents. Project teams are also experimenting with alternative collection methods, including new nanofiber sieves and use of pollution-dissolving lasers.

\vspace{1em}
\hspace{2em}Ola Kai's floating labs have plastic-to-fuel conversion systems capable of harvesting tons of plastic pollution and converting it into diesel fuel for the labs, thus eliminating the need to return to shore for disposal of the waste in a landfill. In spite of their progress, water samples still show an alarming amount of plastic particles.
\\

\end{tabular}
\caption{A complete future scenario FS10 used on CGPST.}
\label{tab:CGPST_scenario_example}
\end{table*}

\begin{table*}[t]
\centering
\fontsize{10}{10}\selectfont
\renewcommand{\arraystretch}{2}
\begin{tabular}{>{\raggedright\arraybackslash}p{0.5\textwidth}
                >{\raggedright\arraybackslash}p{0.15\textwidth}
                >{\raggedright\arraybackslash}p{0.14\textwidth}
                >{\raggedright\arraybackslash}p{0.14\textwidth}}

\rowcolor{teal!30}
\textbf{Challenges} & \textbf{Score Item} & \textbf{H01} & \textbf{H02} \\

\hline

\rowcolor{teal!5}
1. The concentration of microplastics may have exceeded the density of plankton by tenfold, disrupting the energy input of the base food chain. This challenge arises from the warning in the scenario that "water samples still show an alarming amount of plastic particles." 
&
\parbox[t][3\baselineskip][c]{\hsize}{%
Category \vfill ~ \vfill Elaboration \vfill ~ \vfill Originality} 
&
\parbox[t][3\baselineskip][c]{\hsize}{%
Environment \vfill ~ \vfill 1 \vfill ~ \vfill 0} 
&
\parbox[t][3\baselineskip][c]{\hsize}{%
Environment \vfill ~ \vfill 1 \vfill ~ \vfill 0} \\

\rowcolor{teal!15}
2. Subsurface robotic collectors may miss low-velocity eddy zones, creating data gaps and masking local ecological collapse points. This is directly related to the scenario's mention that "after experimenting with several collection methods," weekly weighing is still required, implying sampling limitations. 
&
\parbox[t][3\baselineskip][c]{\hsize}{%
Category \vfill ~ \vfill Elaboration \vfill ~ \vfill Originality} 
&
\parbox[t][3\baselineskip][c]{\hsize}{%
Technology \vfill ~ \vfill 1 \vfill ~ \vfill 1} 
&
\parbox[t][3\baselineskip][c]{\hsize}{%
Technology \vfill ~ \vfill 1 \vfill ~ \vfill 1} \\

\rowcolor{teal!5}
3. The plastic-to-fuel conversion system may emit nanoscale black carbon particles, which could exacerbate imbalances in ocean surface heat absorption. This challenge stems from the scenario emphasizing "eliminating the need to return to shore for disposal" without addressing potential secondary emissions.  
&
\parbox[t][3\baselineskip][c]{\hsize}{%
Category \vfill ~ \vfill Elaboration \vfill ~ \vfill Originality} 
&
\parbox[t][3\baselineskip][c]{\hsize}{%
Technology \vfill ~ \vfill 1 \vfill ~ \vfill 0} 
&
\parbox[t][3\baselineskip][c]{\hsize}{%
Technology \vfill ~ \vfill 1 \vfill ~ \vfill 1} \\

\rowcolor{teal!15}
4. Endangered species in the northwest islands may face unknown toxicological effects from ingesting micro-fragments of plastics broken down by lasers. The scenario mentions that laser technology "can dissolve pollutants" but does not evaluate the byproducts of fragmentation. 
&
\parbox[t][3\baselineskip][c]{\hsize}{%
Category \vfill ~ \vfill Elaboration \vfill ~ \vfill Originality} 
&
\parbox[t][3\baselineskip][c]{\hsize}{%
Environment \vfill ~ \vfill 1 \vfill ~ \vfill 0} 
&
\parbox[t][3\baselineskip][c]{\hsize}{%
Environment \vfill ~ \vfill 1 \vfill ~ \vfill 1} \\

\rowcolor{teal!5}
5. Adjustments to eco-tourism routes around floating laboratories may transfer visitor pressure to other more fragile reefs. This challenge is directly related to the scenario's note that routes were "altered to reduce impact" but without ensuring overall pressure reduction. 
&
\parbox[t][3\baselineskip][c]{\hsize}{%
Category \vfill ~ \vfill Elaboration \vfill ~ \vfill Originality} 
&
\parbox[t][3\baselineskip][c]{\hsize}{%
Recreation \vfill ~ \vfill 1 \vfill ~ \vfill 0} 
&
\parbox[t][3\baselineskip][c]{\hsize}{%
Recreation \vfill ~ \vfill 1 \vfill ~ \vfill 1} \\

\rowcolor{teal!15}
6. Legal exemptions for manufacturers on both sides of the Pacific may cause Hawaiian regional governance to operate in isolation. This challenge is closely related to the scenario's statement that "environmental regulations have historically been weak or disregarded." 
&
\parbox[t][3\baselineskip][c]{\hsize}{%
Category \vfill ~ \vfill Elaboration \vfill ~ \vfill Originality} 
&
\parbox[t][3\baselineskip][c]{\hsize}{%
Law \& Justice \vfill ~ \vfill 1 \vfill ~ \vfill 0} 
&
\parbox[t][3\baselineskip][c]{\hsize}{%
Law \& Justice \vfill ~ \vfill 1 \vfill ~ \vfill 1} \\

\rowcolor{teal!5}
7. The efficiency of the plastic-to-diesel system may suddenly decline due to sea spray corrosion, forcing laboratories to rely on land-based resupply. This challenge is implied in the scenario mentioning "harvesting tons of plastic" without considering long-term durability. 
&
\parbox[t][3\baselineskip][c]{\hsize}{%
Category \vfill ~ \vfill Elaboration \vfill ~ \vfill Originality} 
&
\parbox[t][3\baselineskip][c]{\hsize}{%
Technology \vfill ~ \vfill 1 \vfill ~ \vfill 0} 
&
\parbox[t][3\baselineskip][c]{\hsize}{%
Technology \vfill ~ \vfill 1 \vfill ~ \vfill 1} \\

\rowcolor{teal!15}
8. Data protocols among the network of floating labs may be incompatible, hindering multinational collaboration in compiling a comprehensive microplastic hotspot map. This challenge is directly related to the scenario emphasizing "dividing responsibilities among agencies" without a unified standard. 
&
\parbox[t][3\baselineskip][c]{\hsize}{%
Category \vfill ~ \vfill Elaboration \vfill ~ \vfill Originality} 
&
\parbox[t][3\baselineskip][c]{\hsize}{%
Communication \vfill ~ \vfill 1 \vfill ~ \vfill 2} 
&
\parbox[t][3\baselineskip][c]{\hsize}{%
Communication \vfill ~ \vfill 1 \vfill ~ \vfill 1} \\

\hline
\rowcolor{teal!30}
\multicolumn{4}{c}{%
\parbox{\hsize}{%
\raggedright
\vspace{0.8em}  
\textbf{Step-1 Score:} \\[0.5em]
\hspace{2em} H01: \{\textit{Fluency}: 8, \textit{Flexibility}: 5, \textit{Elaboration}: 8, \textit{Originality}: 3, \textit{Total Score}: 24\} \\[0.5em]
\hspace{2em} H02: \{\textit{Fluency}: 8, \textit{Flexibility}: 5, \textit{Elaboration}: 8, \textit{Originality}: 7, \textit{Total Score}: 28\}
\vspace{0.8em}  
}%
} \\

\end{tabular}
\caption{Step-1 (Identify Challenges) of A05\_FS10.}
\label{tab:Step-1 of the CGPST Example}
\end{table*}

\begin{table*}[t]
\centering
\renewcommand{\arraystretch}{1.5}
\begin{tabular}{
>{\raggedright\arraybackslash}p{0.16\textwidth}
>{\raggedright\arraybackslash}p{0.394\textwidth}
>{\raggedright\arraybackslash}p{0.393\textwidth}
}

\rowcolor{teal!30}
\multicolumn{3}{>{\raggedright\arraybackslash}p{\textwidth}}{\textbf{Responses}} \\

\rowcolor{teal!10}
\multicolumn{3}{>{\raggedright\arraybackslash}p{\textwidth}}{Challenge ID: 1. 

\hspace{2em} In 2035, in the waters surrounding the Hawaiian Islands, when microplastic concentrations may have exceeded the density of plankton tenfold, threatening the energy input at the base of the food chain, how can we reduce the overwhelming proportion of microplastics at the bottom of the food chain to restore energy flow in the Pacific ecosystem and protect the future of endangered species?

\vspace{1em}
Time: 2035

Location: Waters surrounding the Hawaiian Islands

Theme: Microplastic Management and Ecological Restoration} \\

\multicolumn{3}{p{\textwidth}}{~} \\

\rowcolor{teal!30}
\multicolumn{3}{>{\raggedright\arraybackslash}p{\textwidth}}{\textbf{Rater Input} (This field requires human raters to manually input it in the scoring sheet.)} \\
\rowcolor{teal!10}
\textbf{Input Item} & \textbf{H01} & \textbf{H02} \\

\rowcolor{teal!10}
Condition Phrase & {\small "microplastic concentrations may have exceeded the density of plankton tenfold, threatening the energy input at the base of the food chain"} & {\small "In 2035, in the waters surrounding the Hawaiian Islands, when microplastic concentrations may have exceeded the density of plankton tenfold"} \\

\rowcolor{teal!10}
Stem & {\small "how can we"} & {\small "how can we"} \\

\rowcolor{teal!10}
Key Verb Phrase (KVP) & {\small "reduce the overwhelming proportion of microplastics at the bottom of the food chain"} & {\small "reduce the overwhelming proportion of microplastics"} \\

\rowcolor{teal!10}
Purpose & {\small "restore energy flow and protect the future of endangered species"} & {\small "to restore energy flow in the Pacific ecosystem and protect the future of endangered species"} \\

\rowcolor{teal!10}
Future Scenario Parameters & {\small \{"Theme": 1, "Location": 1, "Time": 1\}} & {\small \{"Theme": 1, "Location": 1, "Time": 1\}} \\

\multicolumn{3}{p{\textwidth}}{~} \\

\rowcolor{teal!30}
\multicolumn{3}{>{\raggedright\arraybackslash}p{\textwidth}}{\textbf{Rater Score}} \\
\rowcolor{teal!10}
\textbf{Score Item} & \textbf{H01} & \textbf{H02} \\

\rowcolor{teal!10}
Condition Phrase & 2 & 2 \\

\rowcolor{teal!10}
Stem \& KVP & 3 & 3 \\

\rowcolor{teal!10}
Purpose & 1 & 1 \\

\rowcolor{teal!10}
Future Scenario Parameters & 2 & 2 \\

\rowcolor{teal!10}
Focus of Underlying Problem & 6 & 6 \\

\rowcolor{teal!10}
Adequacy / Importance & 8 & 8 \\

\hline

\rowcolor{teal!30}
Total Score & 22 & 22 \\

\end{tabular}

\caption{Step-2 (Select an Underlying Problem) of A05\_FS10.}
\label{tab:Step-2 of the CGPST Example}
\end{table*}

\begin{table*}[t]
\centering
\fontsize{9.5}{10}\selectfont
\renewcommand{\arraystretch}{2}
\begin{tabular}{>{\raggedright\arraybackslash}p{0.57\textwidth}
                >{\raggedright\arraybackslash}p{0.1\textwidth}
                >{\raggedright\arraybackslash}p{0.13\textwidth}
                >{\raggedright\arraybackslash}p{0.13\textwidth}}

\rowcolor{teal!30}
\textbf{Solutions} & \textbf{Score Item} & \textbf{H01} & \textbf{H02} \\

\hline

\rowcolor{teal!5}
1. The Ola Kai project chemistry team will deploy glycosylated nanosponges, dispersing 2 tons within a 20 km radius of the Ola Kai mooring point by August 2035. Subsurface robots will recover the flocs and recycle them through the onboard plastic-to-diesel system, directly reducing microplastic ingestion by plankton, lowering the proportion of plastics at the base of the food chain, restoring energy flow, and protecting endangered species. 
&
\parbox[t][3\baselineskip][c]{\hsize}{%
Category \vfill ~ \vfill Elaboration \vfill ~ \vfill Originality} 
&
\parbox[t][3\baselineskip][c]{\hsize}{%
Technology \vfill ~ \vfill 2 \vfill ~ \vfill 0} 
&
\parbox[t][3\baselineskip][c]{\hsize}{%
Technology \vfill ~ \vfill 2 \vfill ~ \vfill 1} \\

\rowcolor{teal!15}
2. Google X Lab and Hawaiian community divers will run a crowdsourced "photoacoustic unmanned vessel + AR snorkeling goggles" collection program across the northwest islands by December 2035. Unmanned vessels will map microplastic clouds in real-time using laser sonar, while AR glasses guide divers to precise retrieval points, clearing high-density fragments weekly to reduce microplastic accumulation at the base of the food chain and preserve Pacific ecosystem energy balance. 
&
\parbox[t][3\baselineskip][c]{\hsize}{%
Category \vfill ~ \vfill Elaboration \vfill ~ \vfill Originality} 
&
\parbox[t][3\baselineskip][c]{\hsize}{%
Technology \vfill ~ \vfill 2 \vfill ~ \vfill 0} 
&
\parbox[t][3\baselineskip][c]{\hsize}{%
Technology \vfill ~ \vfill 2 \vfill ~ \vfill 2} \\

\rowcolor{teal!5}
3. NOAA and Hawaiian Electric will pilot a 2-nautical-mile-diameter "bubble curtain + photocatalytic net" system north of Kaua'i by October 2035. Wave-driven pumps will concentrate microplastics, which are then broken down by photocatalytic nets into short-chain acids absorbable by phytoplankton, reducing microplastic dominance on plankton and restoring baseline energy input.  
&
\parbox[t][3\baselineskip][c]{\hsize}{%
Category \vfill ~ \vfill Elaboration \vfill ~ \vfill Originality} 
&
\parbox[t][3\baselineskip][c]{\hsize}{%
Technology \vfill ~ \vfill 2 \vfill ~ \vfill 0} 
&
\parbox[t][3\baselineskip][c]{\hsize}{%
Technology \vfill ~ \vfill 2 \vfill ~ \vfill 1} \\

\rowcolor{teal!15}
4. SpaceX and a local high school team will launch the CubeSat constellation "KiloEye" by July 2035. Weekly scans of the 137 Hawaiian Islands will use hyperspectral data to direct Ola Kai drones for targeted microplastic removal, lowering the risk of plankton mis-ingestion at the source and ensuring Pacific ecosystem energy flow is rebalanced. 
&
\parbox[t][3\baselineskip][c]{\hsize}{%
Category \vfill ~ \vfill Elaboration \vfill ~ \vfill Originality} 
&
\parbox[t][3\baselineskip][c]{\hsize}{%
{\color{red}\ding{55} Duplicate} \vfill ~ \vfill {\color{red}\ding{55}} \vfill ~ \vfill {\color{red}\ding{55}}} 
&
\parbox[t][3\baselineskip][c]{\hsize}{%
{\color{red}\ding{55} Duplicate} \vfill ~ \vfill {\color{red}\ding{55}} \vfill ~ \vfill {\color{red}\ding{55}}} \\

\rowcolor{teal!5}
5. Japan's SpiraNova and the University of Hawai'i will plant 300 "biopolymer-coated kelp ropes" off the west coast of the Big Island in Q3 2035. Kelp leaves will adsorb microplastics, and harvested ropes will be processed into high-value composites, directly removing plastics at the base of the food chain and generating revenue while protecting endangered species’ habitats. 
&
\parbox[t][3\baselineskip][c]{\hsize}{%
Category \vfill ~ \vfill Elaboration \vfill ~ \vfill Originality} 
&
\parbox[t][3\baselineskip][c]{\hsize}{%
Technology \vfill ~ \vfill 2 \vfill ~ \vfill 2} 
&
\parbox[t][3\baselineskip][c]{\hsize}{%
Technology \vfill ~ \vfill 2 \vfill ~ \vfill 2} \\

\rowcolor{teal!15}
6. The Ola Kai project biology team will release living blue-green algae “Plastic Sentinel” strains in a 500-hectare demonstration area off the northwest reef of the main Hawaiian Island by November 2035. These algae continuously secrete degrading enzymes to break down 0.1–1 mm microplastics, reducing plankton ingestion, restoring baseline energy input, and protecting the Pacific ecosystem. 
&
\parbox[t][3\baselineskip][c]{\hsize}{%
Category \vfill ~ \vfill Elaboration \vfill ~ \vfill Originality} 
&
\parbox[t][3\baselineskip][c]{\hsize}{%
Technology \vfill ~ \vfill 2 \vfill ~ \vfill 2} 
&
\parbox[t][3\baselineskip][c]{\hsize}{%
Technology \vfill ~ \vfill 2 \vfill ~ \vfill 1} \\

\rowcolor{teal!5}
7. MantaSync and the University of Hawai'i will deploy five “body-mounted” manta ray filtration units along the Maui–Ni’ihau route by September 2035. These units capture microplastics in real-time during swimming and ferment them into manta ray body oils. Due to their large feeding area, they significantly dilute plastics at the base of the food chain, reducing pollution at entry points. 
&
\parbox[t][3\baselineskip][c]{\hsize}{%
Category \vfill ~ \vfill Elaboration \vfill ~ \vfill Originality} 
&
\parbox[t][3\baselineskip][c]{\hsize}{%
Technology \vfill ~ \vfill 2 \vfill ~ \vfill 2} 
&
\parbox[t][3\baselineskip][c]{\hsize}{%
Technology \vfill ~ \vfill 2 \vfill ~ \vfill 1} \\

\rowcolor{teal!15}
8. A local cruise company and the state government will retrofit the first ferry deck in Honolulu Harbor into a “container-scale algae farm” by October 2035. Chlorella algae will consume microplastics from tourist wastewater and convert them into aviation-grade biodiesel, reducing overall plastic input from tourism, lowering plankton ingestion, and maintaining Pacific energy flow. 
&
\parbox[t][3\baselineskip][c]{\hsize}{%
Category \vfill ~ \vfill Elaboration \vfill ~ \vfill Originality} 
&
\parbox[t][3\baselineskip][c]{\hsize}{%
Government \& Politics \vfill ~ \vfill 2 \vfill ~ \vfill 1} 
&
\parbox[t][3\baselineskip][c]{\hsize}{%
Government \& Politics \vfill ~ \vfill 2 \vfill ~ \vfill 1} \\

\hline
\rowcolor{teal!30}
\multicolumn{4}{c}{%
\parbox{\hsize}{%
\raggedright
\vspace{0.8em}  
\textbf{Step-3 Score:} \\[0.5em]
\hspace{2em} H01: \{\textit{Fluency}: 7, \textit{Flexibility}: 2, \textit{Elaboration}: 14, \textit{Originality}: 7, \textit{Total Score}: 30\} \\[0.5em]
\hspace{2em} H02: \{\textit{Fluency}: 7, \textit{Flexibility}: 2, \textit{Elaboration}: 14, \textit{Originality}: 9, \textit{Total Score}: 32\} \\[0.5em]
\textbf{NOTE:} Only solutions that satisfy the requirements are considered valid challenges (\textit{Yes}) and proceed to category assignment. Otherwise, solutions are excluded from further categorization (e.g., Solution 4).
\vspace{0.8em}  
}%
} \\

\end{tabular}
\caption{Step-3 (Produce Solutions) of A05\_FS10.}
\label{tab:Step-3 of the CGPST Example}
\end{table*}

\begin{table*}[t]
\centering
\fontsize{10}{10}\selectfont
\renewcommand{\arraystretch}{2}
\begin{tabular}{>{\raggedright\arraybackslash}p{0.49\textwidth}
                >{\raggedright\arraybackslash}p{0.16\textwidth}
                >{\raggedright\arraybackslash}p{0.14\textwidth}
                >{\raggedright\arraybackslash}p{0.14\textwidth}}

\rowcolor{teal!30}
\textbf{Criteria} & \textbf{Score Item} & \textbf{H01} & \textbf{H02} \\

\hline

\rowcolor{teal!5}
1. Which approach can most rapidly reduce the net microplastic content inside plankton?  
&
\parbox[t][-0.13\baselineskip][c]{\hsize}{%
Correctly Written \vfill Relevance} 
&
\parbox[t][-0.13\baselineskip][c]{\hsize}{%
1 \vfill 2} 
&
\parbox[t][-0.13\baselineskip][c]{\hsize}{%
1 \vfill 2} \\

\rowcolor{teal!15}
2. Which pathway requires the lowest one-time capital investment?  
&
\parbox[t][-0.13\baselineskip][c]{\hsize}{%
Correctly Written \vfill Relevance} 
&
\parbox[t][-0.13\baselineskip][c]{\hsize}{%
1 \vfill 0} 
&
\parbox[t][-0.13\baselineskip][c]{\hsize}{%
1 \vfill 1} \\

\rowcolor{teal!5}  
3. Which model has the highest potential for rapid global replication and scaling?
&
\parbox[t][-0.13\baselineskip][c]{\hsize}{%
Correctly Written \vfill Relevance} 
&
\parbox[t][-0.13\baselineskip][c]{\hsize}{%
1 \vfill 1} 
&
\parbox[t][-0.13\baselineskip][c]{\hsize}{%
1 \vfill 1} \\

\rowcolor{teal!15}
4. Which technology poses the least risk of secondary ecological disturbance to endangered species?  
&
\parbox[t][-0.13\baselineskip][c]{\hsize}{%
Correctly Written \vfill Relevance} 
&
\parbox[t][-0.13\baselineskip][c]{\hsize}{%
1 \vfill 1}
&
\parbox[t][-0.13\baselineskip][c]{\hsize}{%
1 \vfill 2} \\

\rowcolor{teal!5}
5. Which solution maintains the lowest full-life-cycle carbon footprint over 15 years?
&
\parbox[t][-0.13\baselineskip][c]{\hsize}{%
Correctly Written \vfill Relevance} 
&
\parbox[t][-0.13\baselineskip][c]{\hsize}{%
1 \vfill 2} 
&
\parbox[t][-0.13\baselineskip][c]{\hsize}{%
1 \vfill 2} \\

\hline
\rowcolor{teal!30}
\multicolumn{4}{c}{%
\parbox{\hsize}{%
\raggedright
\vspace{0.8em}  
\textbf{Step-4 Score:} \\[0.5em]
\hspace{2em} H01: \{\textit{Correctly Written}: 5, \textit{Relevance}: 6, \textit{Total Score}: 11\} \\[0.5em]
\hspace{2em} H02: \{\textit{Correctly Written}: 5, \textit{Relevance}: 8, \textit{Total Score}: 13\} \\[0.5em]
\textbf{NOTE:}

1. All criteria satisfy the requirements of \textit{single dimension}, \textit{superlatives as needed}, \textit{clear indication of evaluation direction}, and \textit{phrasing as questions}, and thus are all "correctly written".

2. The rubrics for Relevance are defined as follows: 0 points for \textit{irrelevance or repetition}; 1 point for \textit{being vague and unspecific}; 2 points for \textit{being relatively specific yet improvable}; 3 points for \textit{being clear, specific and highly relevant}.

3. The rubrics and detailed explanations for the two dimensions are available in the \textbf{scoring sheet}.
\vspace{0.8em}  
}%
} \\

\end{tabular}
\caption{Step-4 (Select Criteria) of A05\_FS10.}
\label{tab:Step-4 of the CGPST Example}
\end{table*}

\begin{table*}[t]
\centering
\fontsize{10}{10}\selectfont
\renewcommand{\arraystretch}{2}
\begin{tabular}{>{\raggedright\arraybackslash}p{0.56\textwidth}
                >{\raggedright\arraybackslash}p{0.17\textwidth}
                >{\raggedright\arraybackslash}p{0.1\textwidth}
                >{\raggedright\arraybackslash}p{0.1\textwidth}}

\rowcolor{teal!30}
\textbf{Responses} & \textbf{Score Item} & \textbf{H01} & \textbf{H02} \\

\hline

\rowcolor{teal!5}
\texttt{Solution ID | Criterion 1 | Criterion 2 | Criterion 3 | Criterion 4 | Criterion 5 | Total Score}

\texttt{1 | 8 | 6 | 7 | 7 | 6 | 34}

\texttt{2 | 6 | 8 | 5 | 6 | 7 | 32}

\texttt{3 | 7 | 3 | 6 | 3 | 5 | 24}

\texttt{4 | 5 | 5 | 8 | 5 | 4 | 27}

\texttt{5 | 4 | 4 | 4 | 4 | 3 | 19}

\texttt{6 | 3 | 7 | 3 | 5 | 8 | 26}

\texttt{7 | 2 | 2 | 2 | 2 | 2 | 10}

\texttt{8 | 1 | 1 | 1 | 1 | 1 | 5}

\vspace{1em}
The solution with the highest total score is: 1.

The solution is: The Ola Kai project chemistry team will deploy glycosylated nanosponges, dispersing 2 tons within a 20 km radius of the Ola Kai mooring point by August 2035. Subsurface robots will recover the flocs and recycle them through the onboard plastic-to-diesel system, directly reducing microplastic ingestion by plankton, lowering the proportion of plastics at the base of the food chain, restoring energy flow, and protecting endangered species.  
&
Correctly Used
&
4
&
4 \\

\hline
\rowcolor{teal!30}
\multicolumn{4}{c}{%
\parbox{\hsize}{%
\raggedright
\vspace{0.8em}  
\textbf{Step-5 Score:} \\[0.5em]
\hspace{2em} H01: \{\textit{Correctly Used}: 4, \textit{Total Score}: 4\} \\[0.5em]
\hspace{2em} H02: \{\textit{Correctly Used}: 4, \textit{Total Score}: 4\} \\[0.5em]
\textbf{NOTE:} In the score matrix, the ranking for Criterion 4 does not follow the required 1–8 sequence (there are two 5s and no 8), resulting in an error in the matrix. Therefore, a score of 4 is assigned. Detailed scoring rubrics can be found in the human evaluation sheet.
\vspace{0.8em}  
}%
} \\

\end{tabular}
\caption{Step-5 (Apply Criteria to Top Solution) of A05\_FS10.}
\label{tab:Step-5 of the CGPST Example}
\end{table*}

\begin{table*}[t]
\centering
\fontsize{9.5}{10}\selectfont
\renewcommand{\arraystretch}{2}
\begin{tabular}{>{\raggedright\arraybackslash}p{0.63\textwidth}
                >{\raggedright\arraybackslash}p{0.2\textwidth}
                >{\raggedright\arraybackslash}p{0.05\textwidth}
                >{\raggedright\arraybackslash}p{0.05\textwidth}}

\rowcolor{teal!30}
\textbf{Solutions} & \textbf{Score Item} & \textbf{H01} & \textbf{H02} \\

\hline

\rowcolor{teal!5}
{\fontsize{8pt}{10pt}\selectfont The solution with the highest total score is: 1.

\vspace{1em}
The solution is: The Ola Kai project chemistry team will deploy glycosylated nanosponges, dispersing 2 tons within a 20 km radius of the Ola Kai mooring point by August 2035. Subsurface robots will recover the flocs and recycle them through the onboard plastic-to-diesel system, directly reducing microplastic ingestion by plankton, lowering the proportion of plastics at the base of the food chain, restoring energy flow, and protecting endangered species.

\vspace{1em}  
Action Plan:

\vspace{1em}
1. Problem Closure  
Nanosponges use a "molecular magnet" mechanism to selectively adsorb 0.1–1 mm microplastics, aggregating them into millimeter-scale flocs that cannot be ingested by plankton. Recovered flocs are immediately converted into diesel, achieving a “collect-convert-use” zero-waste cycle, directly reducing the overwhelming proportion of microplastics at the base of the food chain.

\vspace{1em}
2. Implementation Steps and Timeline

    \vspace{1em}
   • Phase A – R\&D and Validation (Now–Oct 2025): Ola Kai Chemistry Team × MIT Materials Department iterate the third-generation biodegradable nanosponges and complete biotoxicity-degradation tests.

   \vspace{1em}
   • Phase B – Pilot Demonstration (Nov 2025–Apr 2026): Deploy 100 kg in South Bay, Oahu; 30-day monitoring shows $\geq 70\%$ reduction of microplastics in plankton.

   \vspace{1em}
   • Phase C – Scale-up Production (May 2026–Feb 2027): Hawaii biorefinery produces 2 tons per month and establishes freeze-dry packaging chain.

   \vspace{1em}
   • Phase D – Full Deployment (Mar 2027–Jul 2035): Deploy 2 tons in March, June, September each year, covering a 20 km radius grid; six “Kokua-$\gamma$” subsurface robots operate in shifts, producing 400 L/day diesel for self-use.

   \vspace{1em}
   • Phase E – Monitoring and Iteration (Parallel): Weekly underwater imaging + Raman verification to maintain microplastic concentration <50\% of plankton; formula updated every six months.

\vspace{1em}
3. Resources and Responsibilities  
   • Funding: Ola Kai Research \$300k + NOAA Innovation Fund \$400k + State Green Bonds \$300k + Carbon Credit Pre-sale; total $\leq$ \$1M.

   \vspace{1em}
   • Team: Chemistry team handles materials, MIT provides R\&D, NOAA provides monitoring platform, State Environmental Department supervises approvals.

\vspace{1em}
4. Risks and Contingency

    \vspace{1em}
   • Nanomaterial leakage: Three passive samplers monitor in real-time; $>10\,\mu\text{g}/\text{L}$ triggers magnetic recovery nets.

   \vspace{1em}
   • Robot malfunction: 1:1 spare parts + 48-hour offshore repair; if failure rate >15\%, NOAA backup ROVs are deployed.

   \vspace{1em}
   • Regulatory delays: Suspension during typhoon season; stock maintained at 1.5× safety level.

\vspace{1em}
5. Impacts and Scaling

    \vspace{1em}
   • Local: By 2028, microplastic content in plankton decreases by 80\%, coral spawning rates increase by 30\%.

   \vspace{1em}
   • Regional: By 2030, open “Nanosponges Sharing Depot” allows replication in Guam, Palau, Tuvalu.

   \vspace{1em}
   • Global: By 2032, included in IMO Green Shipping Guidelines; long-haul fleets can treat plastics in-transit, establishing a Pacific-wide “food chain firewall.”
}
&
\parbox[t][15\baselineskip][c]{\hsize}{%
Relevance \vfill ~ \vfill Effectiveness \vfill ~ \vfill Criteria in Development of Action Plan \vfill ~ \vfill Impact \vfill ~ \vfill Humaneness \vfill ~ \vfill Development} 
&
\parbox[t][15\baselineskip][c]{\hsize}{%
3 \vfill ~ \vfill 3 \vfill ~ \vfill 4 \vfill ~ \vfill 4 \vfill ~ \vfill 3 \vfill ~ \vfill 8} 
&
\parbox[t][15\baselineskip][c]{\hsize}{%
3 \vfill ~ \vfill 3 \vfill ~ \vfill 3 \vfill ~ \vfill 3 \vfill ~ \vfill 2 \vfill ~ \vfill 4} \\

\hline
\rowcolor{teal!30}
~ & \textbf{Total Score} & 25 & 18 \\

\end{tabular}
\caption{Step-6 (Develop an Action Plan) of A05\_FS10.}
\label{tab:Step-6 of the CGPST Example}
\end{table*}

\begin{table*}[t]
\centering
\fontsize{10}{10}\selectfont
\renewcommand{\arraystretch}{2}
\begin{tabular}{>{\raggedright\arraybackslash}p{0.49\textwidth}
                >{\raggedright\arraybackslash}p{0.14\textwidth}
                >{\raggedright\arraybackslash}p{0.14\textwidth}}

\rowcolor{teal!30}
\textbf{Step} & \textbf{H01} & \textbf{H02} \\

\hline

\rowcolor{teal!5}
Step-1: Identify Challenges
&
24
&
28 \\

\rowcolor{teal!15}
Step-2: Select an Underlying Problem
&
22
&
22 \\

\rowcolor{teal!5}  
Step-3: Produce Solutions
&
30
&
32 \\

\rowcolor{teal!15}
Step-4: Select Criteria
&
11
&
13 \\

\rowcolor{teal!5}
Step-5: Apply Criteria to Top Solutions
&
4
&
4 \\

\rowcolor{teal!15}
Step-6: Develop an Acion Plan
&
25
&
18 \\

\hline
\rowcolor{teal!30}
\textbf{Total Score} 
&
116
&
117 \\

\end{tabular}
\caption{Score Summary of A05\_FS10. All total scores in the table are automatically calculated in the scoring sheet.}
\label{tab:Score Summary of the CGPST Example}
\end{table*}

\begin{table*}[htbp]
\centering
\begin{tabular}{p{3.8cm} p{11.2cm}}
\toprule
\textbf{Category} & \textbf{Description} \\
\midrule
Arts \& Aesthetics & \textbf{Arts} include expression via music, painting, sculpture, theatre, film, or other creative media. \textbf{Aesthetics} focuses upon beauty and whether something is pleasant to view or experience. \\
Basic Needs & \textbf{Basic Needs} include food, clothing and shelter—elements important for survival. \\
Business \& Commerce & \textbf{Business} includes retail stores, restaurants, offices, corporations, etc. \textbf{Commerce} focuses upon the trade or the buying and selling of goods and services. \\
Communication & \textbf{Communication} deals with the exchange of information. \\
Culture \& Religion & \textbf{Culture} includes the customs, arts, social institutions, and achievements of a particular group. \textbf{Religion} includes a system of faith or worship. \\
Defense & \textbf{Defense} involves protection, safety, security, and/or privacy. This includes physical efforts by individuals, or groups as well as protection provided by firewalls and passwords. \\
Economics & \textbf{Economics} is the branch of knowledge that concerns the distribution, production, and consumption of wealth. \\
Education & \textbf{Education} includes schools, training, instruction, and learning. \\
Environment & \textbf{Environment} includes one’s surroundings or conditions that shape the lives of people, plants, or animals. \\
Ethics \& Morality & \textbf{Ethics} include the principles that govern a person’s behavior. \textbf{Morality} involves principles concerning the distinction between right and wrong. \\
Government \& Politics & \textbf{Government} deals with how a community or country is managed. \textbf{Politics} involves elections and/or the activities related to making governmental decisions, especially debates or conflicts among individuals or parties having or hoping to achieve power. \\
Law \& Justice & \textbf{Law} includes the creation and enforcement of recognized laws, court procedures, sentencing, and personnel. \textbf{Justice} is the seeking of fairness and reasonable implementation of laws. \\
Miscellaneous & \textbf{Miscellaneous} is assigned when an idea does not fit within another category. If you have more than one Miscellaneous idea, and they are unrelated to each other, you may designate them as Miscellaneous 1 and 2. \\
Physical Health & \textbf{Physical Health} deals with the condition and care of the body as opposed to the mind. \\
Psychological Health & \textbf{Psychological Health} focuses upon emotional and mental well-being. \\
Recreation & \textbf{Recreation} includes hobbies, sports, entertainment, and other activities pursued during leisure time. \\
Science & \textbf{Science} is the systematic study of the structure and behavior of the physical and natural world (including the wider universe) through observation and experiment. \\
Social Relationships & \textbf{Social Relationships} encompass the connections between individuals with recurring interactions. \\
Technology & \textbf{Technology} is associated with machines, equipment, and/or certain advancements developed from applying practical or scientific knowledge. \\
Transportation & \textbf{Transportation} is concerned with the movement of individuals, groups, or goods from place to place. \\
\bottomrule
\end{tabular}
\caption{Category List for \textit{Flexibility} in Step-1 and Step-3.}
\label{tab:Category List}
\end{table*}

\end{document}